\pdfoutput=1
\documentclass[11pt]{article}
\usepackage{amsmath,amsfonts,bm}
\usepackage[]{acl}

\usepackage{times}
\usepackage{latexsym}

\usepackage[T1]{fontenc}
\usepackage[utf8]{inputenc}

\usepackage{microtype}

\usepackage{hyperref}
\usepackage{url}
\usepackage{booktabs}
\usepackage{enumitem}
\usepackage{wrapfig}
\usepackage{pifont}
\usepackage{bbding}
\usepackage{adjustbox}
\usepackage{multirow}
\usepackage{bbding}
\usepackage[normalem]{ulem}
\usepackage{graphicx}
\usepackage[normalem]{ulem}
\usepackage{longtable}
\usepackage{tabularx}
\usepackage{iitem}
\usepackage{soul}
\usepackage{color, xcolor} 
\usepackage{bbm}
\usepackage{comment}
\useunder{\uline}{\ul}{}
\title{\textsc{Knowledge Crosswords}: Geometric Knowledge \\ Reasoning with Large Language Models}

\author{Wenxuan Ding\thanks{\quad equal contribution} \ $^1$ \ \ \ \ \ \ \ \ \  Shangbin Feng\footnotemark[1] \ $^2$ \ \ \ \ \ \ \ \ \  Yuhan Liu$^3$ \ \ \ \ \ \ \ \ \  Zhaoxuan Tan$^4$ \\ \textbf{Vidhisha Balachandran}$^5$ \ \ \ \ \ \ \ \ \ \textbf{Tianxing He}$^2$ \ \ \ \ \ \ \ \ \ \textbf{Yulia Tsvetkov}$^2$ \\
$^1$The Hong Kong University of Science and Technology \ \ \ $^2$University of Washington \\ $^3$Xi'an Jiaotong University \ \ \ $^4$University of Notre Dame \ \ \ $^5$Carnegie Mellon University \\
\texttt{wdingaj@connect.ust.hk}, \texttt{shangbin@cs.washington.edu}}

\newcommand{\ourbenchmark}{{\textsc{Knowledge Crosswords}}}
\newcommand{\stagebystage}{\textsc{Staged Prompting}}
\newcommand{\verifyall}{\textsc{Verify-All}}
\begin{document}
\graphicspath{{images/}}

\maketitle
\begin{abstract}
%Large language models (LLMs) are widely adopted in knowledge-intensive tasks and have achieved impressive performance thanks to their knowledge abilities. While LLMs have demonstrated outstanding performance on atomic or linear QA tasks, whether they can reason in knowledge-rich scenarios with interweaving constraints remains an underexplored problem. In this work, 
%We propose \emph{geometric knowledge reasoning}, a new knowledge-intensive QA setting where incomplete pieces of knowledge are interconnected in a network while language models (LMs) are tasked with inferring the missing facts bounded by geometric constraints. 
We propose {\ourbenchmark}, a geometric knowledge reasoning benchmark consisting of incomplete knowledge networks bounded by structured factual constraints, where LLMs are tasked with inferring the missing facts to meet all constraints. The novel setting of geometric knowledge reasoning necessitates new LM abilities beyond existing atomic/linear multi-hop QA, such as backtracking, verifying facts and constraints, reasoning with uncertainty, and more. {\ourbenchmark} contains 2,101 individual problems, covering diverse knowledge domains, and is further divided into three difficulty levels. We conduct extensive experiments to evaluate existing LLMs and approaches on {\ourbenchmark}. Results demonstrate that baseline approaches struggle with larger knowledge networks and semantically-equivalent entity distractors. In light of their limitations, we propose two new approaches, {\stagebystage} and {\verifyall}, to augment LLMs' abilities for error-aware backtracking and constraint verification. Our {\verifyall} significantly outperforms prior methods and is more robust towards problems in the hard subset. Further analysis shows that geometric knowledge reasoning poses new challenges to LLMs' knowledge abilities, particularly in robustness towards varying option orders, complex structural constraints in knowledge networks, ``none of the above'' scenarios, and more.\footnote{Code and data are publicly available at \href{https://github.com/Wenwen-D/KnowledgeCrosswords}{https://github.com/Wenwen-D/KnowledgeCrosswords}.}
\end{abstract}

\section{Introduction}

\begin{figure*}[t]
    \centering
    \includegraphics[width=1\linewidth]{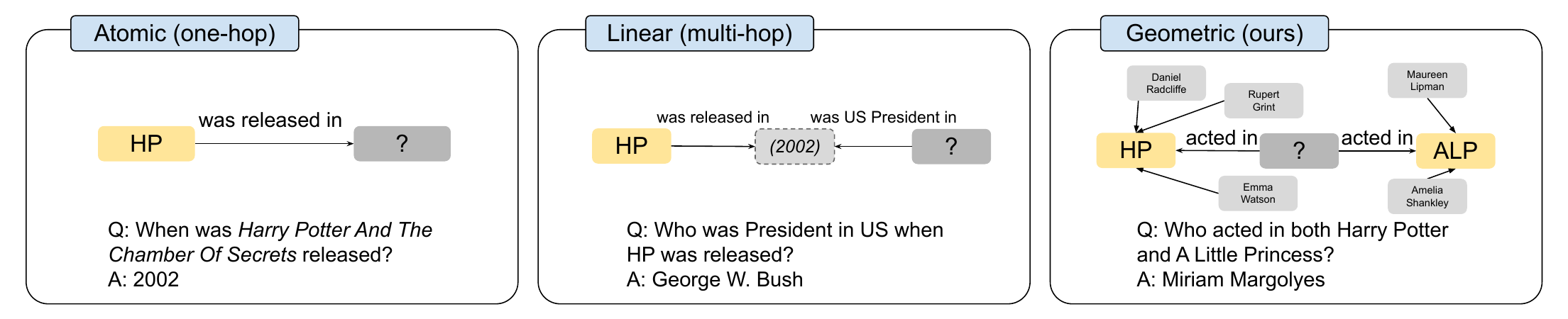}
    % \vspace{-0.1in}
    \caption{Illustration of the differences of atomic, linear (multi-hop), and \emph{geometric knowledge reasoning}. Each step of atomic or linear QA leads to a unique and definite (intermediate) answer, while multiple candidates in each step should be jointly considered to satisfy structural constraints in geometric knowledge reasoning.}
    \label{fig:teaser}
    %\vspace{-0.5in}
\end{figure*}

Large language models (LLMs) encode wast amounts of world knowledge in model parameters \citep{petroni2019language,yu2023kola}.
Existing tasks and datasets assess LLM knowledge abilities mostly by focusing on atomic (e.g., open-domain QA) \citep{rajpurkar2016squad, das2022knowledge,joshi-etal-2017-triviaqa} or linear (e.g., multi-hop QA) \citep{press2022measuring,yang-etal-2018-hotpotqa,ho-etal-2020-constructing} settings, extracting one fact or a fixed concatenation of facts from LLMs. However, knowledge (and language) is naturally structured \citep{reagans2003network}, going beyond linear arrangements, involving complex structural attributes, and forming an interweaving network that connects various entities and relations through multiple chains as illustrated in Figure~\ref{fig:teaser}.

Consequently, we ask: \emph{Can LLMs extend beyond linear compositionality and aggregate information from multiple chains along with various knowledge constraints?} 
Specifically, can LLMs, with the help of their internal parametric knowledge and inherent reasoning patterns, infer missing facts in a network? We term such ability \emph{geometric knowledge reasoning}. 
While compositional QA has been explored in the constrained setting of \emph{external} knowledge bases \citep{zelle1996learning,cui2017kbqa,ye2022rng,neelam2022sygma,xie2022sequential}, we aim to investigate whether LLMs could reason with non-linear fact networks solely relying on their \emph{internal} parametric knowledge.
Formally, we define \emph{geometric knowledge reasoning} as reasoning over a network by inferring missing entities based on the given contextual information, where such networks cannot be simply broken down into chains (like linear reasoning), risking the loss of structural information and constraints. 
% Specifically, when certain pieces of knowledge are missing, can LLMs successfully fill in the blanks based on existing constraints represented by other available information in the network? 
%In this work, we evaluate how well models can aggregate information from the given constraints across a graph representing pieces of knowledge and figure out the blanks in this graph with their inherent reasoning ability.
Geometric knowledge reasoning with LLMs naturally necessitates new LLM abilities beyond those encountered in atomic or linear tasks, such as composing knowledge across multiple chains, reasoning with uncertainty, fact verification, error-aware backtracking, and more.
Since state-of-the-art LLMs are trained on linear sequences of texts devoid of explicit structure, it is underexplored whether they could effectively apply their linearly acquired knowledge to solve geometric reasoning tasks. LLMs with strong geometric knowledge reasoning abilities could serve as versatile relational databases, allowing controllable access to parametric knowledge through SQL-like conditioned prompting.

% While compositional knowledge QA has been explored in the setting of \emph{knowledge bases} , to what extent could \emph{LLMs} exhibit these abilities and handle contexts of geometric knowledge reasoning remain underexplored. 

To this end, we propose {\ourbenchmark}, a geometric knowledge reasoning dataset with 2,101 problems evaluating to what extent LLMs could achieve such desiderata. Each knowledge crossword consists of a list of constraints representing an incomplete fact network, and LLMs need to reconstruct the knowledge network while ensuring that all factual constraints are met. To solve knowledge crosswords, LLMs should ideally evaluate candidates for each blank, jointly consider factual constraints, verify intermediate solutions, and backtrack when encountering factual errors, until all constraints are met.
%We build individual knowledge crosswords by sampling subgraphs from an encyclopedic knowledge graph and randomly masking out certain entities as blanks.
For each incomplete fact network, we generate three sets of distractors as options that are progressively more plausible, resulting in \emph{easy}, \emph{medium}, and \emph{hard} subsets for fine-grained evaluation. Each problem also comes with two settings: \emph{w/o knowledge}, where LLMs solve knowledge crosswords solely with parametric knowledge; \emph{w/ knowledge}, where a helpful (and noisy) paragraph is prepended to each problem.
%In total, {\ourbenchmark} contains 2,101 problems, covering varying levels of difficulty, knowledge domains, and more.

% To sample such subgraphs, we first obtain the neighborhood of a randomly selected center node with a moderate degree, then downsample it to a desired network size by removing nodes that are presumably less informative. The blanks are subsequently picked from nodes with relatively high degrees in the subgraph so that LLMs can have more constraints to work with. We additionally sample optional relevant knowledge for each knowledge crossword problem which could be prepended to the question description to serve as a hint. For each knowledge crossword, we generate negative samples of three difficulty levels for each blank to result in a multiple-choice setting. In total, {\ourbenchmark} contains 2,101 problems, covering varying levels of difficulty, knowledge domains, and more.
% sets of knowledge crossword constraints, 359 of them have options on a difficult level, 873 of them have options on a medium level and 870 of them have options on an easy level.

We conduct extensive experiments to evaluate LLMs and approaches on {\ourbenchmark}, ranging from simple zero-shot prompting to advanced ones such as self-consistency (\textsc{SC}) \citep{wang2022self} and least-to-most prompting (\textsc{LtM}) \citep{zhou2022least}. Results demonstrate that baselines struggle with problems in the hard subset and have significant performance drops in the \emph{w/o knowledge} setting. Advanced prompting methods such as \textsc{SC} and \textsc{LtM} barely improve LLMs due to their reliance on left-to-right reasoning patterns. To address these challenges, we propose two new instruction-based techniques, {\stagebystage} and {\verifyall}, aiming at augmenting LLMs' abilities for backtracking, constraint verification, and more.
{\stagebystage} guides LLMs through an elaborate problem-solving process that progressively solves and simplifies the problem blank by blank, while {\verifyall} proposes candidates for all blanks and verifies them simultaneously.
We find that {\verifyall} achieves top performance and is more robust with harder problems, while the success of {\stagebystage} is contingent on stronger base LLMs. Further analysis reveals geometric knowledge reasoning poses great challenges to LLM knowledge abilities, as they could be susceptible to a variety of factors such as option order, ``none of the above'' scenarios, number of distractors, special structural patterns, and more. We envision geometric knowledge reasoning as a challenging research question and {\ourbenchmark} as a comprehensive testbed to evaluate LLM knowledge abilities in more complex and structured settings.

\begin{figure*}[t]
    \centering
    \includegraphics[width=1\linewidth]{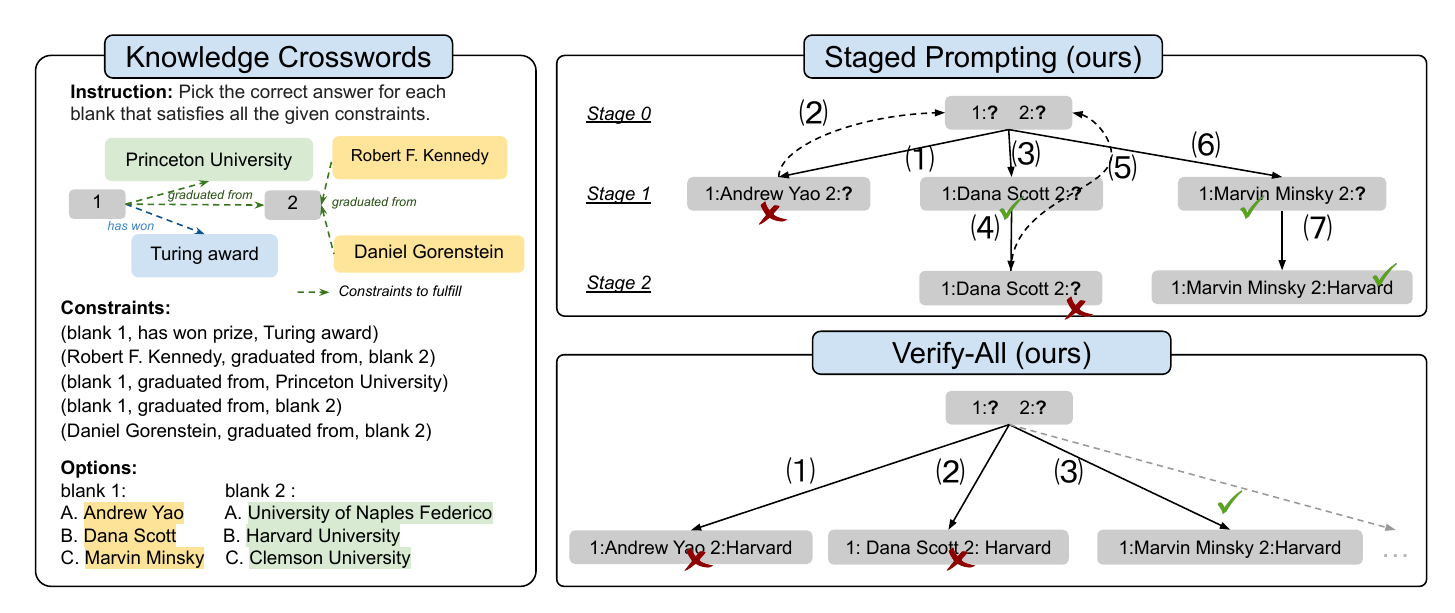}
    % \vspace{-0.1in}
    \caption{Overview of {\ourbenchmark} and two proposed approaches, {\stagebystage} and {\verifyall}. 
    Each knowledge crossword includes task instruction, factual constraints, and multiple-choice QA options. In each stage of {\stagebystage}, LLMs \emph{\ding{172} solve} one blank based on one remaining constraint; \emph{\ding{173} update} the status by filling in the proposed answer; then \emph{\ding{174} verify} filled constraints to proceed or backtrack. In {\verifyall}, LLMs propose a combination of \emph{\ding{172} candidates} and \emph{\ding{173} verify} all constraints with those candidates, and repeat this process until all constraints are met.}
    \label{fig:overview}
    % \vspace{-0.2in}
\end{figure*}

\section{\ourbenchmark} %todo
\label{sec:benchmark}

We propose {\ourbenchmark}, a geometric knowledge reasoning benchmark to evaluate whether LLMs could reason with incomplete fact networks bounded by geometric constraints (Appendix~\ref{sec:benchmark-details}). An example knowledge crossword is presented in Figure~\ref{fig:overview}.

% \paravs
\paragraph{Definition} Each knowledge crossword consists of a question graph $\mathcal{G}_\mathcal{Q} = \{(h,r,t)|h,t \in \mathcal{V_{Q}}, r \in \mathcal{R}\}$, where $\mathcal{V_{Q}}$ is the set of entities represented as nodes of $\mathcal{G}_\mathcal{Q}$ and $\mathcal{R}$ is the set of all possible relations between entities.
%, and $h$, $t$ refer to the head and tail of directed edge in $G_Q$, $r$ refers to the relation of the edge.
Each $(h,r,t)$ in $\mathcal{G}_\mathcal{Q}$ denotes a directed edge representing a factual association such as \textit{(Marvin Minsky, has won prize, Turing award)}. In the question graph $\mathcal{G}_\mathcal{Q}$, certain nodes $b_i \in \mathcal{V_{Q}}$ are masked out as blanks $\mathcal{B} = [b_1, b_2,\ldots,b_{|\mathcal{B}|}]$ for QA. The goal of each knowledge crossword is to find one combination of answers for all blanks $\mathcal{A} = [a_1, a_2,\ldots, a_{|\mathcal{B}|}]$ that satisfies all factual associations represented as geometric constraints in the question graph $\mathcal{G}_\mathcal{Q}$.

% \paravs
\paragraph{Data Source}
We resort to encyclopedic knowledge graphs, specifically YAGO \citep{suchanek2023integrating}, as scaffolds of geometric knowledge reasoning to construct the {\ourbenchmark} benchmark. Different from existing KBQA datasets (ComplexWebQuestions, \citet{talmor-berant-2018-web}, GrailQA, \citet{10.1145/3442381.3449992}, \textit{inter alia}) where LMs are required to reason with \emph{external} KBs, LLMs solve knowledge crosswords with their \emph{internal} parametric knowledge. We conduct preprocessing to remove certain relations in YAGO that are location-related, time-sensitive, or not self-evident. This is to ensure that the {\ourbenchmark} is minimally affected by question ambiguity \citep{min-etal-2020-ambigqa,cole2023selectively}, outdated knowledge \citep{yu2023self,hernandez2023measuring}, etc. We obtain the filtered knowledge graph as $\mathcal{KG}=\{(h,r,t)|h \in \mathcal{H}, r \in \mathcal{R}, t \in \mathcal{T}\}$, where $\mathcal{H}$, $\mathcal{R}$, and $\mathcal{T}$ are the sets of heads, relations, and tails respectively.
% We proceed to use the filtered YAGO as the scaffold to generate knowledge crosswords.
%To be more specific, QGs with location-related edges tend to result in multiple answers, which may confuse language models and make evaluation difficult. Time-sensitive edges can be factually incorrect if not bounded by a certain time range. And the edges that are not self-evident are unclear without detailed definitions. 
% Add a graph to demonstrate?
% \paravs
\paragraph{Question Graphs}
We first adopt two hyperparameters to control the property and difficulty of the generated question graphs (incomplete fact networks): \emph{question graph size} $s_G$, denoting the total number of nodes in a question graph, and \emph{blank size} $s_B$, representing the number of nodes masked out as blanks that need to be filled.
We start from a random center node $c$ and retrieve the $k$-hop neighborhood of $c$ as $\mathcal{G}_\mathcal{N}^{(c)}$. We then downsample $\mathcal{G}_\mathcal{N}^{(c)}$ by randomly removing nodes with degrees higher than a dynamic threshold $t_d$ in $\mathcal{KG}$, until the largest weakly connected component in $\mathcal{G}_\mathcal{N}^{(c)}$ has a size no greater than $s_G$. This is motivated by the fact that entities with higher degrees are presumably less typical and more ambiguous, resulting in problem ambiguity \citep{shomer2023toward,qian2023merge}. We refer to the largest connected component in downsampled $\mathcal{G}_\mathcal{N}^{(c)}$ as an answer graph $\mathcal{G}_\mathcal{A}$.

We then randomly select $s_B$ nodes in $\mathcal{G}_\mathcal{A}$ with degrees larger than a threshold $t_b$ in $\mathcal{G}_\mathcal{A}$ and mask them out as blanks $B$ to obtain a question graph. These high-degree blanks would be rich in geometric associations and provide more constraints to work with. The question graph is then linearized in triplet format, since converting interconnected triples into plain natural language and vice versa are noisy and prone to introduce errors and biases \citep{bai2023kgquiz, min2023factscore}.
%The constraints are expressed in a triplet format, where all components are pre-processed to resemble natural language and enhance comprehensibility. Due to the inherent graph structure of the problems, it is impractical to precisely linearize the triplets as natural language.
By employing hyperparameters and thresholds such as $s_G$ and $s_B$, the dataset comes with built-in difficulty control measures to controllably generate diversified problems. As the question graph generation step does not guarantee answer uniqueness, we exhaustively search answers for each $\mathcal{G}_\mathcal{Q}$ in $\mathcal{KG}$ and only retain those with one valid answer combination.

% \paravs
\paragraph{Negative Sampling}
We mainly consider {\ourbenchmark} in a multiple-choice setting, where several candidates are provided for each blank in $\mathcal{G}_\mathcal{Q}$. This would require negative sampling for each blank to provide distractors in addition to the correct answer, while we identify a taxonomy of three progressive rules for distractors from loose to strict:

\begin{itemize}[leftmargin=*]
    \item Rule \#1: \textbf{Semantic Role}: If a blank $b_i$ is the head (or tail) of an edge with relation $r_i$, then the distractor for $b_i$ should be selected from the set of heads (or tails) of edges with the same $r_i$.
    \item Rule \#2: \textbf{Network Proximity}: The distractor for $b_i$ should occur in the neighborhood $\mathcal{G}_\mathcal{N}^{(c)}$ around $c$, which further ensures that distractors are likely to be in a similar context as $b_i$.
    \item Rule \#3: \textbf{Definite Constraint}: If the other end of the edge that the blank $b_i$ is incident to is known, then we say such edge is a definite constraint for $b_i$, and the distractor should satisfy at least one of such definite constraints for $b_i$ to fulfill Rule \#3. Such distractors impose higher demands on LMs in the sense that LMs should jointly consider all constraints to exclude these distractors.
\end{itemize}

% We set three criteria to generate negative samples for each blank $B_i$ of $G_Q$ to control the difficulty: 1. has same sematic role as the blank; 2. exists in the neighborhood $G_N$ around the center node $C$ used to generate $G_Q$; 3. satisfies at least one definite constraints of the blank. For the first criterion, if a blank $B_i$ is the head (or tail) of an edge with relation $r_i$, then the negative sample for $B_i$ could be selected from the set of heads (or tails) of edges with the same relation $r_i$. The second criterion further narrows the range of such edges so that the negative samples are more likely to be in a similar context as the blank. For the third criterion, if the other end of the edge that the blank $B_i$ is incident to is known, then we say such edge is a definite constraint for $B_i$, and the negative sample should satisfy at least one of such definite constraints for $B_i$ to fulfill the third criterion. Such $NS$s makes greater demands from language models in the sense that LMs should jointly consider all constraints to exclude such $NS$s.

As a result, we obtain three negative sampling strategies with varying difficulty implications for knowledge crosswords: \emph{easy}, where distractors only meet Rule \#1; \emph{medium}, where distractors meet Rule \#1 and \#2; \emph{hard}, where distractors meet Rule \#1, \#2, and \#3. We opt to separately assign multiple options for each blank in $\mathcal{G}_\mathcal{Q}$, using either \emph{easy}, \emph{medium}, or \emph{hard} strategies, resulting in three subsets of knowledge crosswords with increasing difficulty levels.
\paragraph{Relevant Knowledge}
Each knowledge crossword comes with two settings: \textsc{w/o knowledge}, where LLMs solely rely on internal parametric knowledge to solve knowledge crosswords; \textsc{w/ knowledge} where a helpful but noisy passage is prepended to the problem description. These knowledge passages contain both helpful information about the correct answers and irrelevant information generated by the three proposed negative sampling rules. Useful and irrelevant knowledge is then sampled to 1:3, shuffled, and presented before each knowledge crossword. LLMs would need to dynamically select relevant and useful information to facilitate geometric knowledge reasoning, which poses new challenges to LLMs \citep{shi2023large}.
%Concretely, we similarly sample confounding knowledge triples from $\mathcal{KG}$, guided by the three rules, to obtain confounding context for each $\mathcal{G}_\mathcal{Q}$.  In this way, we provide both necessary knowledge and confounding information, making the solving process non-trivial.

%To mitigate the potential influence of the lack of relevant knowledge of language models, we propose providing triplets relevant to the constraints in $G_Q$ as one of the experiment settings. For each constraint in $G_Q$, we present four pertinent triplets, with one specifically designated for the correct answer. And the other three are randomly sampled following similar method as negative sampling, with priority given to the triplets that satisfy all the criteria. By sampling relevant triplets in such a way, we provide necessary knowledge (filled with correct answers) as well as confounding knowledge that makes the solving process non-trivial. The sampling of confounding knowledge also simulates the possible information that one may consider when solving the question, with those satisfying all three criteria having the highest likelihood of being considered intuitively.
% \paravs
\paragraph{Evaluation Metrics}
We evaluate performance on {\ourbenchmark} with two metrics: \emph{Partial-Credit} (\textsc{PC}), indicating the portion of blanks that have been answered correctly; \emph{Full-Credit} (\textsc{FC}), indicating whether all blanks are correct in a given knowledge crossword. Formally,
\begin{align*}
\textsc{PC} = \frac{\sum_{i=1}^{s_B}\mathbbm{1}[a_i'=a_i]}{s_B}, \ \ \
\textsc{FC} = \mathbbm{1}[\textsc{PC}=1]
\end{align*}
where $a_i'$ denotes the prediction of blank $b_i$ by LLMs and $\mathbbm{1}[\cdot]$ denotes the indicator function.
% \paravs

\paragraph{Benchmark Statistics}
We obtain 873 valid question graphs with different levels of scales and sparsity. Each question graph is then used to construct three problems using the three levels of negative sampling difficulty, \emph{easy}, \emph{medium}, and \emph{hard}, resulting in a total of 2,101 problems and enabling fine-grained evaluation. The problems are described in English and the benchmark statistics are shown in Table~\ref{tab:mc_statistics}.
\begin{table}[t]
\centering
% \vspace{-10pt}
\resizebox{1\linewidth}{!}{
\begin{tabular}{@{}l c c c c@{}}
\toprule[1.5pt]
          \textbf{Subset} & \textbf{\#Qs} & \textbf{Avg. \#Nodes} & \textbf{Avg. \#Edges} & \textbf{Avg. \#Blanks} \\ \midrule[0.75pt]
\textsc{easy}      & 869  & 7.28      & 6.63       & 2.89       \\
\textsc{medium}    & 873  & 7.28      & 6.64       & 2.89       \\
\textsc{hard} & 359  & 7.09      & 6.41       & 2.62       \\ \bottomrule[1.5pt]
\end{tabular}
}
% \vspace{-10pt}
\caption{Statistics of the {\ourbenchmark} Benchmark. We report the number of questions and the average number of nodes, edges, and blanks for each subset with different negative sampling difficulty.}
\label{tab:mc_statistics}
\end{table} 

\section{Experiment Settings}
\begin{table*}
\centering
 \resizebox{1\linewidth}{!}{
\begin{tabular}{@{}lcccccccccccc@{}}
\toprule[1.5pt]
 &
  \multicolumn{6}{c}{\textsc{w/ knowledge}} &
  \multicolumn{6}{c}{\textsc{w/o knowledge}} \\
  \cmidrule(lr){2-7} \cmidrule(lr){8-13}
\multirow{3}{*}{\textbf{Method}} &
  \multicolumn{2}{c}{\textbf{easy}} &
  \multicolumn{2}{c}{\textbf{medium}} &
  \multicolumn{2}{c}{\textbf{hard}} &
  \multicolumn{2}{c}{\textbf{easy}} &
  \multicolumn{2}{c}{\textbf{medium}} &
  \multicolumn{2}{c}{\textbf{hard}} \\
  \cmidrule(lr){2-3} \cmidrule(lr){4-5}  \cmidrule(lr){6-7}  \cmidrule(lr){8-9}  \cmidrule(lr){10-11}  \cmidrule(lr){12-13}
  &
  \textsc{PC} &
  \textsc{FC} &
  \textsc{PC} &
  \textsc{FC} &
  \textsc{PC} &
  \textsc{FC} &
  \textsc{PC} &
  \textsc{FC} &
  \textsc{PC} &
  \textsc{FC} &
  \textsc{PC} &
  \textsc{FC} \\ \midrule[0.75pt]
\textsc{Random} &
  34.3 &
  6.1 &
  34.2 &
  5.5 &
  33.5 &
  8.4 &
  34.3 &
  6.1 &
  34.2 &
  5.5 &
  33.5 &
  8.4 \\
\textsc{Upperbound} &
  98.8 &
  96.7 &
  99.1 &
  97.4 &
  91.8 &
  82.2 &
  - &
  - &
  - &
  - &
  - &
  - \\ \midrule[0.75pt]
%\multicolumn{13}{l}{\textit{Experiments with \textsc{gpt-3.5-turbo}} }\\
\textsc{Zero-Shot} &
  97.3 &
  {\ul 93.7} &
  97.4 &
  {\ul 94.2} &
  77.9 &
  55.4 &
  81.3 &
  57.1 &
  83.3 &
  60.6 &
  {\ul 57.2} &
  24.8 \\
\textsc{Few-Shot} &
  {\ul 97.8} &
  93.2 &
  {\ul 97.6} &
  93.5 &
  {\ul 78.8} &
  54.0 &
  {\ul 83.7} &
  {\ul 60.8} &
  {\ul 84.7} &
  {\ul 63.3} &
  56.8 &
  25.3 \\
\textsc{CoT} &
  94.6 &
  86.5 &
  95.0 &
  88.9 &
  77.9 &
  56.3 &
  74.0 &
  44.0 &
  76.4 &
  48.5 &
  55.7 &
  27.0 \\
\textsc{CoT+SC} &
  95.9 &
  89.8 &
  96.6 &
  91.2 &
  78.7 &
  {\ul 57.4} &
  75.2 &
  45.8 &
  77.3 &
  49.1 &
  56.7 &
  {\ul 28.4} \\
\textsc{LtM} &
  86.0 &
  68.9 &
  86.3 &
  68.6 &
  69.6 &
  43.5 &
  75.6 &
  47.3 &
  76.6 &
  48.2 &
  51.1 &
  19.2 \\ \midrule[0.75pt]
\stagebystage &
  91.9 &
  81.6 &
  91.2 &
  80.4 &
  70.5 &
  44.5 &
  64.3 &
  34.3 &
  67.4 &
  38.3 &
  47.9 &
  15.8 \\
\verifyall &
  \textbf{98.6} &
  \textbf{96.1} &
  \textbf{98.7} &
  \textbf{96.2} &
  \textbf{83.9} &
  \textbf{64.6} &
  \textbf{84.5} &
  \textbf{62.3} &
  \textbf{86.1} &
  \textbf{66.9} &
  \textbf{59.7} &
  \textbf{29.8} \\ \midrule[0.75pt]
  \multicolumn{13}{l}{\textit{Experiments with \textsc{gpt-4}} }\\ 
\stagebystage &
  \multicolumn{1}{l}{\textbf{99.1}} &
  \multicolumn{1}{l}{\textbf{98.8}} &
  \multicolumn{1}{l}{\textbf{96.3}} &
  \multicolumn{1}{l}{95.6} &
  \multicolumn{1}{r}{\textbf{95.4}} &
  \textbf{94.2} &
  \multicolumn{1}{l}{75.4} &
  \multicolumn{1}{l}{70.7} &
  \multicolumn{1}{l}{78.8} &
  \multicolumn{1}{l}{74.0} &
  \multicolumn{1}{l}{52.3} &
  \multicolumn{1}{l}{32.4} \\ 
\verifyall &
  \multicolumn{1}{r}{98.1} &
  \multicolumn{1}{r}{98.1} &
  \multicolumn{1}{r}{95.7} &
  \multicolumn{1}{r}{\textbf{95.7}} &
  \multicolumn{1}{r}{92.8} &
  90.5 &
  \multicolumn{1}{r}{\textbf{88.0}} &
  \multicolumn{1}{r}{\textbf{83.4}} &
  \multicolumn{1}{r}{\textbf{89.5}} &
  \multicolumn{1}{r}{\textbf{85.5}} &
  \multicolumn{1}{r}{\textbf{59.5}} &
  \multicolumn{1}{r}{\textbf{38.6}} \\
  % \midrule \midrule
% \multicolumn{13}{l}{\textit{Experiments with \textsc{gpt-4}} }\\ 
% \stagebystage &
%   \multicolumn{1}{l}{78.8*} &
%   \multicolumn{1}{l}{75.3*} &
%   \multicolumn{1}{l}{82.7*} &
%   \multicolumn{1}{l}{79.2*} &
%   \multicolumn{1}{l}{54.7*} &
%   \multicolumn{1}{l}{34.7*} &
%   \multicolumn{1}{l}{} &
%   \multicolumn{1}{l}{} &
%   \multicolumn{1}{l}{97.7} &
%   \multicolumn{1}{l}{97.5} &
%   \multicolumn{1}{r}{96.4} &
%   95.2 \\ 
% \verifyall &
% \multicolumn{1}{r}{87.1} &
%   \multicolumn{1}{r}{82.0} &
%   \multicolumn{1}{r}{86.9} &
%   \multicolumn{1}{r}{82.2} &
%   \multicolumn{1}{r}{58.6} &
%   \multicolumn{1}{r}{37.9} &
%   \multicolumn{1}{r}{96.3} &
%   \multicolumn{1}{r}{96.3} &
%   \multicolumn{1}{r}{95.8} &
%   \multicolumn{1}{r}{95.8} &
%   \multicolumn{1}{r}{92.8} &
%   90.5 \\
  \bottomrule[1.5pt]
\end{tabular}
}
\caption{FC and PC with \textsc{gpt-3.5-turbo} unless otherwise specified. The best results are \textbf{bold-faced}, and the second-best ones are \underline{underlined}. Notably, {\verifyall} outperforms the second-best baselines by 7.2\% and 1.4\% (FC) on the hard subset under \textsc{w/ knowledge} and \textsc{w/o knowledge} respectively.
% [one sentence about how verify-all outperforms baselines by at least ?\% on hard subsets]
}
\label{tab:main-results}
\end{table*}

\subsection{Baselines}
\label{sec:baselines}
We evaluate various prompting approaches on {\ourbenchmark}, including Zero-Shot (\textsc{Zero-Shot}) prompting, Few-Shot in-context learning (\textsc{Few-Shot}), Chain-of-Thought prompting (\textsc{CoT}), CoT with Self-Consistency (\textsc{CoT+SC}), and Least-to-Most prompting (\textsc{LtM}). Besides, we adopt the \textsc{Random} baseline which refers to randomly selecting an option for each blank. We also present an \textsc{Upperbound} baseline, where we present oracle knowledge to the LLM, \emph{i.e.} the constraints in $G_Q$ filled with correct answers.

\subsection{Models and Settings}
Unless otherwise specified, we use ChatGPT (\textsc{gpt-3.5-turbo}) as the base language model in our experiments, and we additionally test out GPT-4 and open-source models such as Llama 2 \citep{touvron2023llama}. For Few-Shot prompting techniques (\textsc{Few-Shot}, \textsc{CoT}, \textsc{CoT+SC}, \textsc{LtM}), we present 5 in-context exemplars. The sampling temperature $\tau$ is set to $0.1$ except for \textsc{CoT+SC}; we sample $5$ Chain-of-Thought responses with temperature $\tau = 0.7$ for the CoT with Self-Consistency baseline.

\section{Our Approach}

We hypothesize that the left-to-right reasoning patterns in autoregressive language models \citep{yao2023tree} and prompting approaches (discussed in section \ref{sec:baselines}) would fall short of solving knowledge crosswords, which require backtracking, maintaining problem states, verifying existing information, reasoning with structured constraints, and more. To this end, we introduce two instruction-based methods that promote these abilities, illustrated with a detailed example in Figure~\ref{fig:overview}.
% \subsecvs
\subsection{\stagebystage}
The {\stagebystage} approach divides geometric knowledge reasoning into stages involving three steps: \emph{solve}, \emph{update}, and \emph{verify}. At the beginning of each stage, LLMs maintain a current status of solved blanks and unresolved constraints (edges that involve unsolved blanks). In the \emph{solve} step, LLMs propose a candidate for an unsolved blank based on internal knowledge; in the \emph{update} step, LLMs update unsolved constraints using the newly proposed candidate for an associated blank; in the \emph{verify} step, LLMs reflect on the updated constraints in the \emph{update} step and judge their validity. If an invalid factual association is identified as a result of the proposed candidate, LLMs backtrack to the problem status in the previous stage and propose another option; otherwise, LLMs proceed to tackle the remaining blanks until all blanks are filled and all constraints are met.

% \subsecvs
\subsection{\verifyall}
While {\stagebystage} presents an elaborate problem-solving process that tackles challenges such as backtracking and status updates, such complex reasoning might be hard to learn in context for LLMs. We additionally propose the {\verifyall} approach: candidates for each blank are simultaneously proposed, rather than in separate stages. A verification step is then employed to assess the validity of all filled constraints using these proposed candidates. If an error is detected, the LM should backtrack and propose an alternative set of candidates until no error is found.
%\uline{Verify-all prompting example, code: explain here or in results?}

\section{Results}
We evaluate approaches on {\ourbenchmark} and present results in Table~\ref{tab:main-results}.
% \paravs
\paragraph{LLMs have preliminary abilities for geometric knowledge reasoning.} Table~\ref{tab:main-results} shows that all investigated approaches outperform the \textsc{Random} baseline, while LLMs could achieve 90+ FC scores on the \emph{easy} subset and \emph{w/ knowledge} setting. However, model performance (FC) drops by 29.7\% on average on the \emph{hard} subset compared to \emph{easy}, even when the required knowledge stays the same, indicating that LLMs are far from robust on geometric knowledge reasoning.
% \paravs
% \paragraph{Performance degrades with stronger distractors.} An average Full-Credit drop of 33.4\% and 25.9\% is observed from the \emph{easy} subset to the \emph{hard} under the \textsc{w/ knowledge} and \textsc{w/o knowledge} respectively, indicating that LLM abilities for geometric knowledge reasoning are not solely contingent on its internal parametric knowledge and factuality, susceptible to distractors and irrelevant context.
%While it might be relatively straightforward for LLMs to eliminate false options that satisfy few definite constraints in the knowledge crosswords, they face difficulties in excluding options that meet a subset of all the constraints presented. This indicates that LLMs' abilities for geometric knowledge reasoning are greatly impacted by how confounding distractors are, meaning that LLMs are far from robust in complex structured knowledge contexts.
% \paravs
\paragraph{Noisy knowledge does help LLMs solve knowledge crosswords.} Despite the existence of irrelevant and confounding information, LLMs do benefit from the prepended noisy knowledge. On average, the \textsc{w/ knowledge} settings exhibit a 34.3\% \textsc{FC} gain compared to \textsc{w/o knowledge} settings across all approaches. This indicates that LLMs possess preliminary abilities to selectively leverage knowledge and information.
%However, it remains unclear whether LLMs obtain such performance gain by effectively understanding and leveraging the relevant knowledge or by spurious correlations between relevant knowledge and given constraints.
% \paravs
\paragraph{Advanced prompting methods show little improvement.} Specifically, \textsc{CoT}, \textsc{LtM} and \textsc{CoT+SC} do not greatly advance performance compared to \textsc{Zero-Shot} and \textsc{Few-Shot} prompting: the average \textsc{PC} and \textsc{FC} of \textsc{CoT}, \textsc{LtM} and \textsc{CoT+SC} is 5.6\% and 10.4\% \textbf{less} than those of \textsc{Zero-Shot} and \textsc{Few-Shot}. This suggests that the left-to-right reasoning patterns employed by these prompting techniques may not be applicable for knowledge crosswords, as these prompting methods fail to induce non-linear reasoning steps for verification and backtracking.
% \paravs
\begin{figure}[t]
    \centering
    \includegraphics[width=1\linewidth]{ 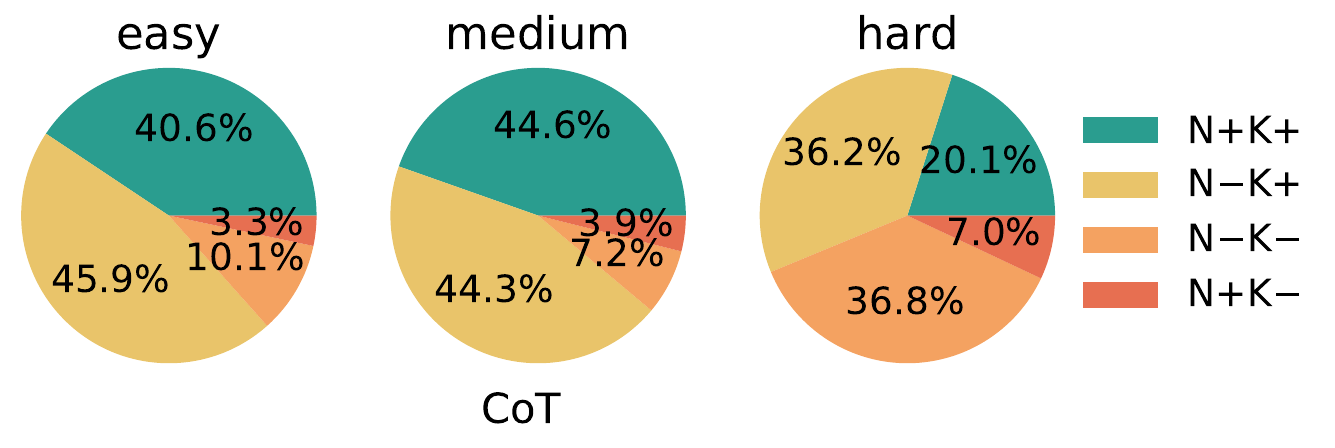}
    \includegraphics[width=1\linewidth]{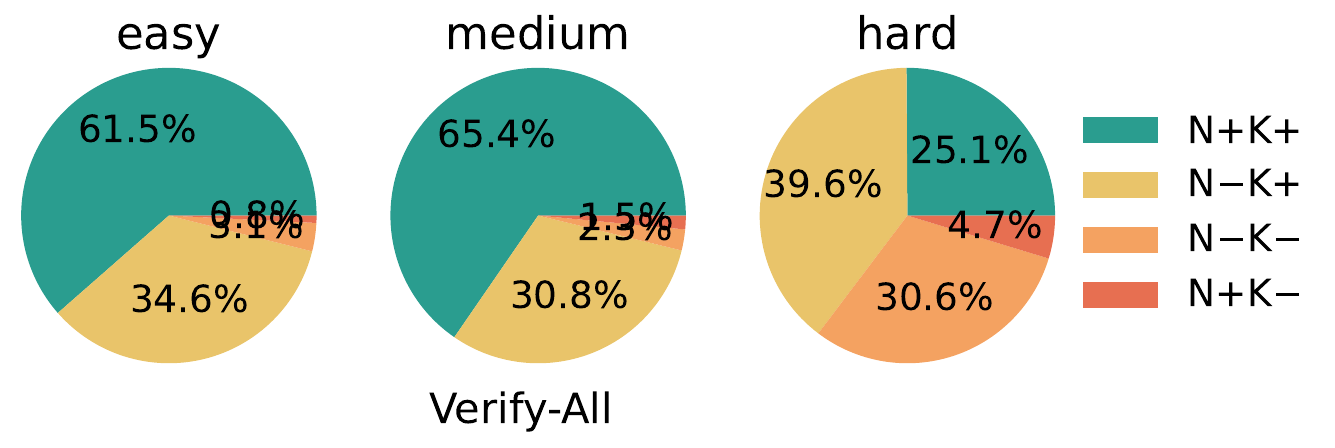}
    \caption{Problem distribution based on the correctness under the \textsc{w/ knowledge} and \textsc{w/o knowledge} settings using \textsc{CoT} and \verifyall. The results indicate that while easier problems are mainly hindered by a lack of knowledge, the bottleneck for hard problems lies in geometric knowledge reasoning abilities.}
    \label{fig:error-analysis}
\end{figure}
\paragraph{Promoting verification and backtracking improves geometric knowledge reasoning.} With \textsc{gpt-3.5-turbo}, {\verifyall} greatly outperforms all baselines with explicit self-verification. Interestingly, after a closer look into the responses, we find that factual errors are rarely detected while the performance gain mainly comes from LLMs proposing more precise answers in a single attempt when specifically asking for fact verification. In addition, while \textsc{gpt-3.5-turbo} struggles at learning complex reasoning steps required by {\stagebystage}, Table~\ref{tab:main-results} shows that {\stagebystage} achieves impressive results with \textsc{gpt-4} and generally outperforms all other methods including {\verifyall} in the \textsc{w/ knowledge} setting. This indicates that the more elaborate instructions of {\stagebystage} work best with more advanced LLMs, as smaller models struggle to grasp these detailed reasoning steps.

% \footnotetext{Within 4k-context, in the w/o Relevant Knowledge setting, the numbers of finished responses for easy, medium, and hard questions are 755, 781, and 341 respectively; in the w/ Relevant Knowledge setting, the numbers of finished responses for easy, medium, hard questions are 759, 769 and 328 respectively. The credits are calculated based on these finished responses only.}

\section{Analysis}
\label{sec:analysis}
\vspace{10pt}
\paragraph{Error Analysis}
Each knowledge crossword comes with \textsc{w/} and \textsc{w/o knowledge} settings. We conduct error analysis to investigate the impact of noisy passages on LLM problem solving and present in Figure \ref{fig:error-analysis}. We label each problem based on whether it is answered (in)correctly in \textsc{w/ knowledge} (K+(--)) and answered (in)correctly in \textsc{w/o knowledge} (N+(--)).
Figure~\ref{fig:error-analysis} reveals that for easy and medium problems, the main bottleneck is knowledge access since most N-- questions are correctly answered under \textsc{w/ knowledge}. However, for hard problems, the bottleneck is geometric knowledge reasoning abilities, given that the proportion of N--K-- is consistently above 30\%, showing that LLMs struggle to reason even when the required knowledge is provided. By including three subsets of varying difficulty in {\ourbenchmark}, we successfully reveal the multitudes of LLM limitations in knowledge-intensive contexts, disentangling limitations of knowledge and reasoning.

%type \#4 problems are rare across all difficulty levels and approaches, suggesting that given knowledge is mostly consistent with the model's parametric knowledge and model is seldom distracted by the confounders. Excluding the hard level, type \#2 constitutes a significantly larger portion than type \#3, indicating the model's preliminary geometric reasoning ability when provided with relevant knowledge for easier problems. However, for hard problems, the main gap lies in weak geometric reasoning ability rather than lack of relevant knowledge as the percentage of type \#3 increases compared to easier problems.

% Please add the following required packages to your document preamble:
% \usepackage{booktabs}
% \usepackage{multirow}
% \usepackage{graphicx}

\vspace{10pt}
\paragraph{None of the Above}
\begin{table}[t]
\centering
% \vspace{0.2in}
\resizebox{1\linewidth}{!}{
\begin{tabular}{@{}cccccccc@{}}
\toprule[1.5pt]
\multirow{2}{*}{\textbf{w/ NOTA?}}&\multirow{2}{*}{\textbf{w/ correct?}} & \multicolumn{2}{c}{\textbf{easy}}& \multicolumn{2}{c}{\textbf{medium}} & \multicolumn{2}{c}{\textbf{hard}} \\
\cmidrule(lr){3-4} \cmidrule(lr){5-6} \cmidrule(lr){7-8}
& & \textsc{PC}           & \textsc{FC}          & \textsc{PC}         & \textsc{FC}         & \textsc{PC}          & \textsc{FC}         \\ \midrule[0.75pt]
\checkmark & \XSolidBrush & 35.7 & 14.0 & 35.6 & 13.5 & 11.4 & 3.1  \\
\checkmark & \checkmark  & 
{\ul 68.1}& {\ul 46.8} & {\ul 69.1} & {\ul 48.3} & {\ul 50.7} & {\ul 20.6} \\
\XSolidBrush & \checkmark & \textbf{83.7} & \textbf{60.8} & \textbf{84.7} & \textbf{63.3} & \textbf{56.8} & \textbf{25.3} \\ \bottomrule[1.5pt]
\end{tabular}
}
\caption{FC and PC (\%) with \textsc{Few-Shot} in the \textsc{w/o knowledge} setting. The best results are \textbf{bold-faced} and the second-best \underline{underlined}. ``w/ NOTA'' denotes where LMs are asked to consider none-of-the-above through instructions and ``w/ correct'' denotes whether the correct combination is actually provided.}
\label{tab:NOTA}
% \vspace{-0.2in}
\end{table}

%In the main experiments, the correct answer is always provided as one of the options for each blank and LLMs are instructed to pick the correct answer. 
To study whether LLMs are subject to ``none-of-the-above (NOTA) scenarios'', we add an instruction \textit{``Output `none of the above' if none of the option combinations satisfy all the constraints.''} and evaluate the performance of \textsc{gpt-3.5-turbo} with the \textsc{Few-shot} prompting. 
Specifically, we experiment with two settings: \#1. The correct option is removed from candidates and LLMs should choose to report NOTA; \#2. The correct option exists and LLMs should provide answers.
Table~\ref{tab:NOTA} demonstrates that LLMs struggle with the NOTA scenario, regardless of whether the knowledge crossword is indeed coming without correct answers.
%seldom outputs ``None of the above'' even if the correct answer is not provided as one option (row 1 \& row 2), while it does get affected by the absence of the assumption that correct option always exists (row 2 \& row 3).

% \paravs
\begin{figure}[t]
    \centering
    \includegraphics[width=1\linewidth]{ 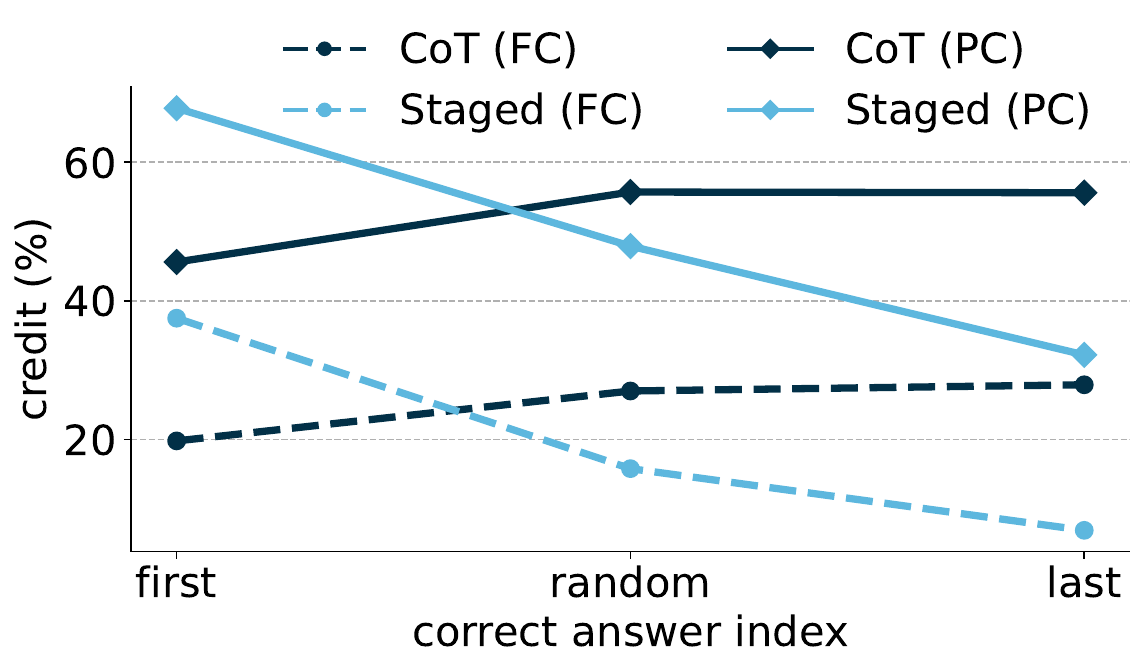}
    \caption{FC and PC (\%) under the \textsc{w/o knowledge} setting using \textsc{CoT} and {\stagebystage} with different orders of options evaluated on the hard subset.}
    \label{fig:order}
    % \vspace{10pt}
\end{figure}
\paragraph{Option Order}
As our proposed {\stagebystage} considers one candidate at one time, we expect that model performance may be worse for problems where the correct answer appears later in the prompt. Figure~\ref{fig:order} demonstrates this negative correlation, which could be attributed to LLM hallucination \citep{ji2023survey} and falsely accepting earlier incorrect options. On the other hand, we see an opposite trend in the performance of \textsc{CoT}. This indicates that \textsc{CoT} does not consider the options sequentially and later options might influence the prediction more.

% \paravs

% \begin{figure}[t]
%     \centering    \includegraphics[width=0.8\linewidth]{bar.pdf}
%     \vspace{-0.2in}
%     \caption{Model performance (\%) for problems with specific structural patterns. 
%     ``overall'' denotes the performance calculated over the whole benchmark. 
%     % \emph{A-B} denotes problems with two blanks connected together. \emph{A-B-C} denotes problems with three problems connected together. \emph{cycle} denotes problems where the blanks form a cycle in the question graph.
%     }
%     \label{fig:pattern}
%     \vspace{-0.2in}
% \end{figure}

\paragraph{Structural Patterns}
\begin{figure}[t]
    \centering
    % \vspace{-0.2in}
    \includegraphics[width=1\linewidth]{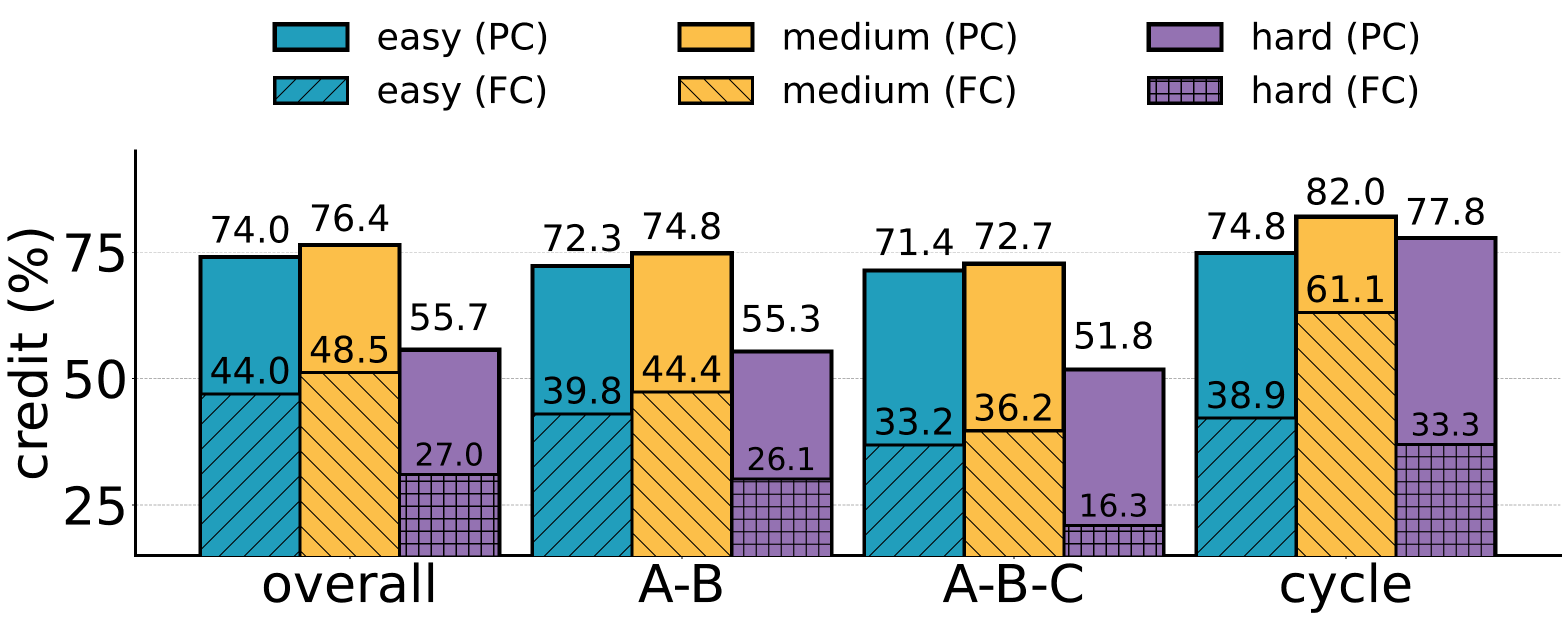}
    \caption{FC and PC (\%) for problems with specific structural patterns using \textsc{CoT}. ``A-B'' denotes two connected blanks, ``A-B-C'' denotes a chain of three blanks, ``cycle`` denotes a cycle of three or more blanks, and
    ``overall'' denotes the performance on all question graphs.
    % \emph{A-B} denotes problems with two blanks connected together. \emph{A-B-C} denotes problems with three problems connected together. \emph{cycle} denotes problems where the blanks form a cycle in the question graph.
    }
    % \vspace{10pt}
    \label{fig:pattern}
\end{figure}

We investigate whether special structural patterns of blanks in the question graphs might impact LLM performance. We identify three patterns: 1) \emph{A-B}, where two blanks are connected by an edge; 2) \emph{A-B-C}, where three blanks are on a chain; 3) \emph{cycle}, where three more more blanks form a cycle. Figure~\ref{fig:pattern} demonstrates a decrease in performance of \emph{A-B} and \emph{A-B-C} compared to the full dataset, showing a chain of blanks would pose challenges to LLM knowledge reasoning. On the other hand, \emph{cycle} exhibits performance gains: we hypothesize that it has a higher blank-to-constraint ratio (closer to 1:1) than other patterns, which gives LLMs more constraints to work with.
% Though the reported performance of problems consisting of a cycle of blanks is above average, this gain may be attributed to the lack of such examples.

\paragraph{Number of In-Context Exemplars}

\begin{figure}[t]
    \centering
    % \vspace{-0.1in}
    \includegraphics[width=1\linewidth]{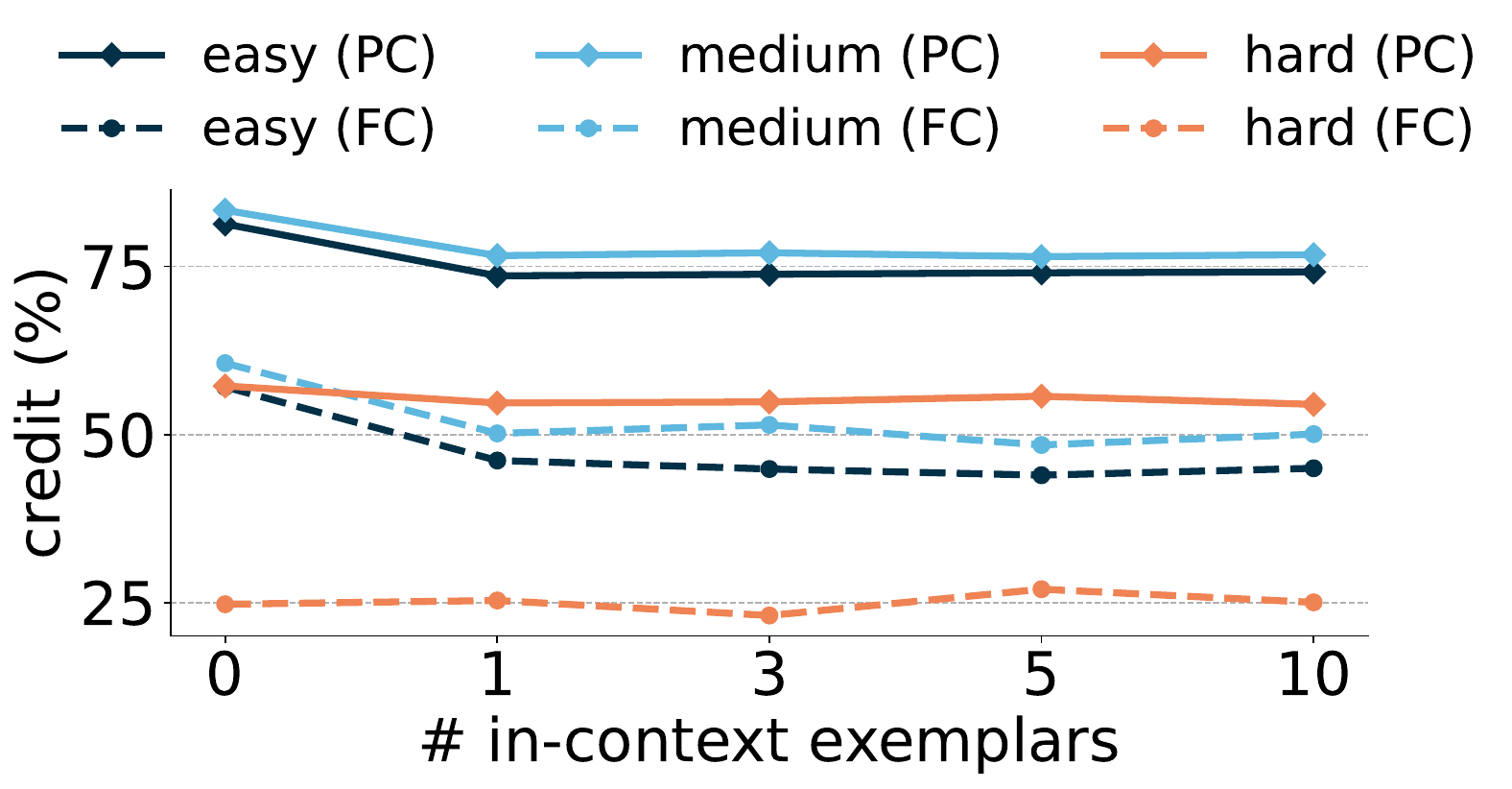}
    % \vspace{-0.2in}
    \caption{FC and PC (\%) using \textsc{CoT} under the \textsc{w/o knowledge} setting with an increasing number of in-context exemplars. An increase in the number of exemplars does not necessarily bring performance gain.}
    % \vspace{10pt}
    \label{fig:nICL}
    % \vspace{-0.2in}
\end{figure}

Despite the in-context learning ability demonstrated by LLMs \citep{brown2020language}, we find that more in-context exemplars fail to improve model performance on \ourbenchmark. As presented in Figure~\ref{fig:nICL}, for questions with all three difficulty levels, the best performance is achieved at \textsc{Zero-Shot} except for the Full-Credit of hard problems. This indicates that left-to-right CoT reasoning could not be adequately learned in context for the problem of knowledge crosswords. 

% \vspace{-0.2in}
% \paravs
\paragraph{Difficulty of In-Context Exemplars}

\begin{table}[t]
\centering
% \vspace{-0.1in}
\resizebox{1\linewidth}{!}{
\begin{tabular}{@{}ccccccc@{}}
\toprule[1.5pt]
&
  \multicolumn{6}{c}{\textbf{test}}\\ \cmidrule{2-7}
\multirow{3}{*}{\textbf{exemplar}} & \multicolumn{2}{c}{\textbf{easy}}& \multicolumn{2}{c}{\textbf{medium}} & \multicolumn{2}{c}{\textbf{hard}} \\
\cmidrule(lr){2-3} \cmidrule(lr){4-5} \cmidrule(lr){6-7}
 & \textsc{PC}           & \textsc{FC}          & \textsc{PC}         & \textsc{FC}         & \textsc{PC}          & \textsc{FC}         \\ \midrule[0.75pt]
 \textbf{easy}    & 73.9 & 44.4 & 75.9 & 47.9 & 55.2 & 24.0  \\
 \textbf{medium}  & \textbf{75.4} & \textbf{47.9} & \textbf{77.2} & \textbf{49.7} & 55.1 & 25.9  \\
 \textbf{hard}    & 73.4 & 43.5 & 76.2 & 48.7 & 55.3 & 24.0  \\ 
\textbf{mixed}    & 74.9 & 46.0 & 75.7 & 47.4 & \textbf{55.8} & \textbf{26.5} \\ 
 \bottomrule[1.5pt]
\end{tabular}
}
\caption{FC (\%) with \textsc{CoT} using exemplars of varying difficulties under the \textsc{w/o knowledge} setting. The best results are in \textbf{bold}.}
\label{tab:diffICL}
% \vspace{10pt}
\end{table}

We investigate the correlation between the difficulty of in-context exemplars and model performance by evaluating the performance with 5-shot \textsc{CoT} using 4 different sets of in-context exemplars: \emph{easy}, where all in-context examples come from the easy subset; similarly \emph{medium} and \emph{hard}; \emph{mixed}, where a mixture of 2 easy, 2 medium, and 1 hard examples are employed.
%While these exemplars share the same question graphs, their distractors are sampled based on rules of different difficulty levels. 
Table~\ref{tab:diffICL} demonstrates that medium or mixed in-context examples work best, while solely employing easy or hard ones is marginally worse. As a result, we follow the \emph{mixed} settings in the main experiments.
%for problems of different difficulties, LLMs learn better from medium or mixed in-context exemplars. This indicates that LLMs best conduct geometric knowledge reasoning when problems and solutions with a range of difficulty levels are presented, enabling progressive in-context learning from simple to hard. As a result, we use in-context exemplars with mixed difficulties in the experiments.

% \paragraph{More Analysis}
% Due to space limit, additional analysis is presented in Appendix \ref{sec:add-analysis}.

\paragraph{Fine-tuning and open-source LMs}
We additionally evaluate the geometric knowledge reasoning abilities of an open-source language model - \textsc{Llama2-7b} with 100 problems randomly selected across all difficulty subsets. Without fine-tuning, \textsc{Llama2-7b} demonstrates a performance close to random guess. After instruction-tuning with 1,471 knowledge crosswords randomly selected from all 2,101 questions, the Partial-Credit and Full-Credit become 17.7\% and 12.0\% higher than \textsc{Zero-Shot} prompting as reported in Table~\ref{tab:finetune}. This indicates that instruction tuning \citep{wei2021finetuned} could augment LLMs for solving knowledge crosswords, while to what extent they work with larger LLMs requires further research.
\begin{table}[t]
\centering
% \vspace{-0.2in}
\begin{tabular}{@{}lcc@{}}
\toprule[1.5pt]
\textbf{Method}       & \multicolumn{1}{l}{\textsc{PC}} & \multicolumn{1}{l}{\textsc{FC}} \\ \midrule[0.75pt]
\textsc{Random}             & 32.8                   & 5.0                      \\
\textsc{Zero-Shot}             & 29.6                   & 5.0                      \\
\textsc{Few-Shot}           & 35.5                   & 7.0                      \\
\textsc{instruction-tuning} & 47.3                   & 17.0                     \\ \bottomrule[1.5pt]
\end{tabular}

\caption{
% Performance (\%) of \textsc{Llama2-7b} on 100 randomly sampled problems. 
\textsc{Zero-Shot} and \textsc{Few-Shot} results are evaluated with \textsc{Llama2-7b} on 100 randomly sampled problems without fine-tuning. After \textsc{instruction-tuning} on 1,471 knowledge crosswords, the performance improves. 
}
\label{tab:finetune}
% \vspace{-0.2in}
% \vspace{-0.1in} 
% \vspace{13pt}
\end{table}

\paragraph{Number of Options}

\begin{figure}[t]
    \centering
    % \vspace{-0.13in}  
\includegraphics[width=1\linewidth]{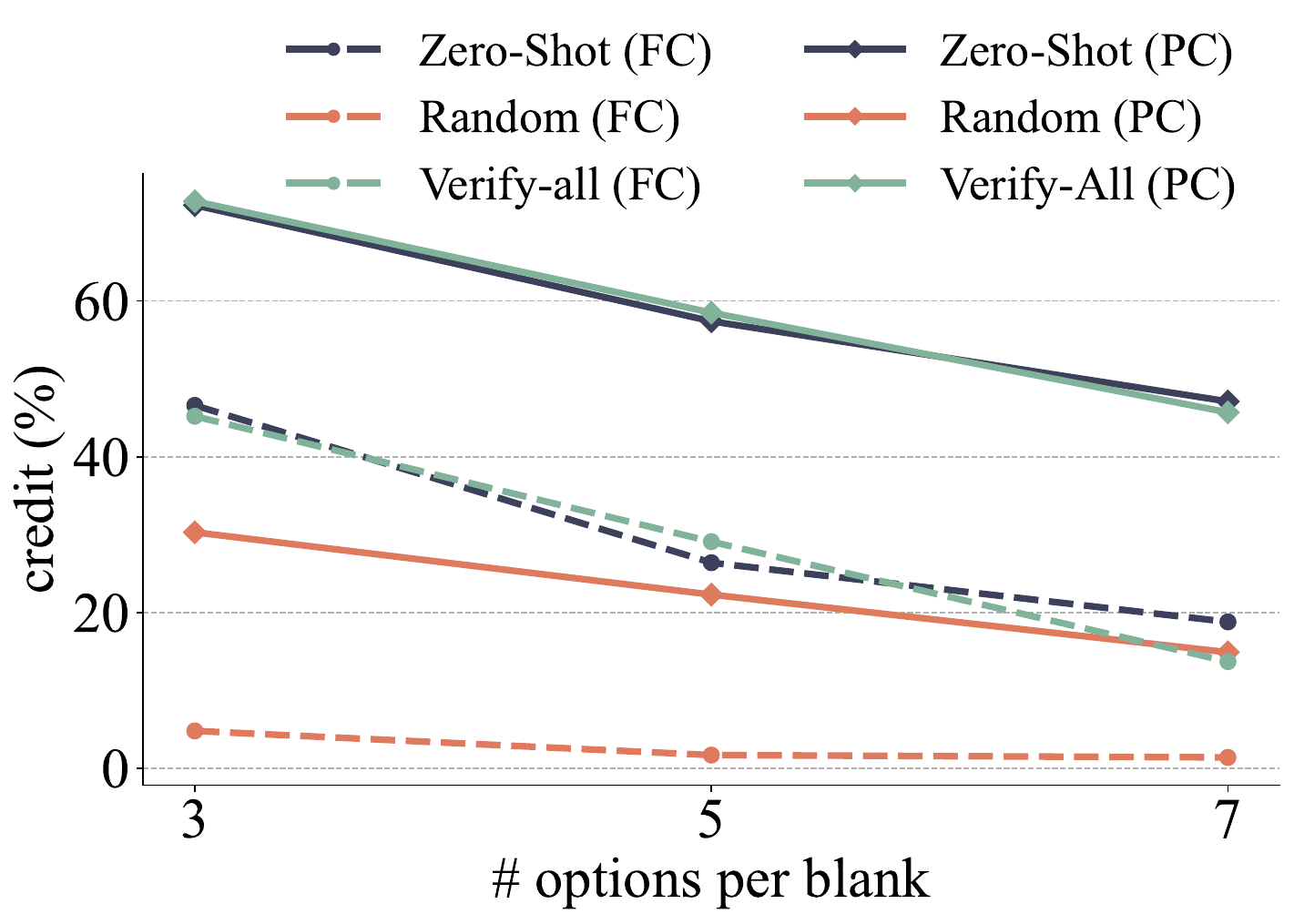}
    \caption{FC and PC (\%) evaluated on 292 problems using \textsc{Zero-Shot} for increasing number of options per blank. \textsc{Random} denotes the baseline of random guess.}
    \label{fig:n-options}
    % \vspace{-0.25in}
\end{figure}

As the number of options for each blank increases, the problem becomes harder due to the presence of more confounders. We expect to see a downward trend in model performance when there are more distractors per blank. Unsurprisingly, the results in Figure~\ref{fig:n-options} show that the performance is negatively correlated with the number of options per blank. We also observe that performance gap with random guessing is narrowing, suggesting that {\ourbenchmark} might be difficult for LLMs in the \textsc{w/o knowledge} generation setting. (Appendix~\ref{sec:discussion})
\section{Related Work}
% \secvsbelow
% \paravs
% \vspace{10pt}
\paragraph{Understanding LLM Knowledge} 
% [understanding]

LLMs could memorize and encode factual knowledge in model parameters \citep{petroni2019language, yu2023kola}. As a result, previous research focuses on investigating the extent to which LLMs retrieve and utilize factual knowledge \citep{yu2022generate, chen2023beyond, mallen2023not}. However, the knowledge abilities of LLMs also come with a wide range of limitations such as knowledge update \citep{hase2023methods}, irrelevant information \citep{shi2023large}, and more \citep{chen2022rich, mruthyunjaya2023rethinking, kandpal2023large, sun2023head,kandpal2023large,xie2023adaptive,amayuelas2023knowledge,wang2023resolving,huang2023large}. As a result, various lines of research aim to expand the knowledge abilities of language models, such as better prompting \citep{press2022measuring,sun2022recitation,yu2022generate,kojima2022large,ye2022unreliability}, retrieval augmentation \citep{shi2023replug,yu2023chain,asai2023self,borgeaud2022improving,jiang2023active}, search engine integration \citep{yu2022generate,press2022measuring,kasai2022realtime,qin-etal-2023-webcpm,vu2023freshllms,khalifa-etal-2023-shot}, and more. While these works primarily focus on evaluating and improving abilities to handle atomic (e.g., open-domain QA) or linear (e.g., multi-hop QA) knowledge, we propose to assess whether LLMs could reason with fact networks bounded by geometric constraints that better align with the structural nature of knowledge. 
%show that while LLMs possess the potential to recall factual information, their ability to capture complex topological and semantic traits
%of KGs remains notably limited. In addition, LLMs encounter challenges related to knowledge update , knowledge conflict \citep{}, irrelevant context , long-tail knowledge \citep{} and more.

%[expanding] 

\vspace{10pt}
\paragraph{Reasoning over Knowledge Graphs}
Simple and complex questions in KBQA have been extensively studied, covering varied tasks including temporal QA \citep{li2022complex,shang2022improving,chen2023multi,saxena2022sequence,xia2022metatkg,mei2022adaptive,ding2022semantic,kannen2022targeted}, conversational QA \citep{ke2022knowledge}, general QA \citep{zhang2022subgraph,bai2022squire}, and more. A myriad of methods have been proposed to tackle these problems, including enhancing reasoning over knowledge graphs \citep{cao2022program,cao2022kqa,ye2022rng,neelam2022sygma,xie2022sequential,patidar-etal-2023-knowledge,zhang-etal-2023-fc,gupta2018neural,zettlemoyer2009learning,cui2017kbqa, zhong2017seq2sql,shen2019multi} and incorporating the generating ability of language models \citep{liu2022uni,shu2022tiara,tang2022improving,hu2022empowering,zhang2022drlk,jiang2023reasoninglm,wang-etal-2023-query,kim-etal-2023-factkg,li-etal-2023-shot,guo2023parse,aglionby-teufel-2022-faithful,zhang2022greaselm}.
% existing approaches primarily rely on linguistic \citep{gupta2018neural,zettlemoyer2009learning} or deep learning-based \citep{cui2017kbqa, zhong2017seq2sql,shen2019multi} semantic parsing to extract logic forms and retrieve answers from knowledge bases. 
However, it remains underexplored whether LLMs, with the help of its \emph{internal} parametric knowledge, could perform geometric knowledge reasoning with elements similar to these works. We propose {\ourbenchmark} to investigate LLMs' ability to utilize their linearly acquired knowledge for structured knowledge reasoning scenarios.
%the extent to which LLMs can handle geometric reasoning remains an underexplored question, and knowledge crosswords go beyond complex QA which can be represented as logic rules to cover a broader range of problems.
% \paravs
\vspace{10pt}
\paragraph{Reasoning with Large Language Models} 
% \paragraph{LLMs for Few-Shot reasoning.}  
LLMs have been evaluated on a myriad of reasoning tasks in an in-context learning setting, including math problems \citep{ling-etal-2017-program, lewkowycz2022solving}, logical reasoning \citep{srivastava2023beyond, huang2022few}, factual knowledge reasoning \citep{press2022measuring,feng2024don}, commonsense reasoning \citep{talmor-etal-2019-commonsenseqa,fang2024complex,wang2023car,wang2023cat,wang2024candle}, and more. 
Leveraging the in-context learning behavior of LLMs, various prompting techniques \citep{wei2022chain,zhou2022least,khot2022decomposed,wang2022self,wang-etal-2023-plan,schick2023toolformer,gao2023pal} have been proposed to boost the reasoning ability. Specifically, \citet{khot2022decomposed} and \citet{yao2023tree} incorporate programs as guides to LLM generation. 
% While \citet{zhang2022greaselm} improves model reasoning with graph neural networks, extending it to problems with more than two hops is challenging due to the over-smoothing issue and limited interpretability. 
In this work, we focus on the geometric knowledge reasoning ability of LLMs, which is different from existing left-to-right reasoning patterns, with minimal explicit program-based guidance, involving self-verification, backtracking, and more.

\vspace{10pt}
\section{Conclusion}
\vspace{10pt}
% \secvsbelow
We propose {\ourbenchmark} to investigate LLMs for geometric knowledge reasoning, \emph{i.e.} inferring missing information from an incomplete fact network bounded by geometric constraints. Extensive experiments demonstrate that while existing prompting approaches struggle to solve problems in the \emph{hard} subset, our proposed {\stagebystage} and {\verifyall} strategies advance model performance while augmenting LLMs with abilities to verify facts, backtrack, and more. Further analysis reveals that LLMs are brittle to ``none-of-the-above'' scenarios, challenging structural patterns, spurious correlations such as option order, and more.
%We envision {\ourbenchmark} as a comprehensive testbed to evaluate LLM knowledge abilities in advanced graph-based contexts.
%a multi-blank QA dataset where each problem consists of a natural language question representing the geometric constraints of an incomplete entity network and LLMs are tasked with working out the missing entities while meeting all factual constraints.  We adopt a variety of prompting methods on {\ourbenchmark} and find that all investigated approaches perform above random, while we further propose {\stagebystage} and {\verifyall} to tackle unique challenges in geometric knowledge reasoning such as backtracking and fact verification. As a result, {\verifyall} outperforms all other techniques with ChatGPT while {\stagebystage} works better with the more advanced GPT-4. Further analysis shows that the geometric reasoning ability of LLMs over structured knowledge is far from robust, as it is impacted by factors such as the order of options, ``None of the above'' scenarios, certain structure patterns entailing greater uncertainty, and more. 
% \newpage
\section*{Limitations}
\paragraph{Limited Data Source} We construct {\ourbenchmark} based on only the encyclopedic knowledge graph YAGO, which covers topics on general knowledge about people, cities, countries, movies, and organizations from Wikidata. Since we will make the code publicly available, we leave it to future work on evaluating the geometric reasoning ability of LLMs on different topics with various data sources, such as biomedical knowledge graphs \citep{chang2020benchmark} and networks in social sciences \citep{feng2022political}.

\paragraph{Answer Uniqueness} Due to the incompleteness of knowledge graphs, it is possible that the answer to a problem in {\ourbenchmark} is not unique. However, such likelihood is presumably low and does not hurt the general evaluation of the geometric reasoning ability of LLMs.

\section*{Acknowledgements}
This material is based upon work supported by the National Science Foundation under CAREER Grant No.~IIS2142739 and NSF Grant No.~IIS2203097. We also gratefully acknowledge support from Alfred P.~Sloan Foundation Fellowship.

\bibliographystyle{acl_natbib}
\bibliography{custom}

\vspace{10pt}

\newpage

\appendix

\section{Discussion}
\label{sec:discussion}
\paragraph{Geometric Reasoning in the \textsc{w/o knowledge} generation setting} While we mainly focus on solving knowledge crosswords in a multiple-choice setting, it is interesting to evaluate the geometric reasoning ability in the \textsc{w/o knowledge} generation setting. Specifically, the problems in {\ourbenchmark} have unique answers, which should be useful when switching to the \textsc{w/o knowledge} generation setting as answer uniqueness makes evaluation easier and makes the problem clearer. Our preliminary experiments show that solving knowledge crosswords in the \textsc{w/o knowledge} generation setting is much harder. Considering the model performance in the multiple-choice setting, one method that might be promising is to prompt LLMs themselves to generate candidates for each blank and thereby transform the \textsc{w/o knowledge} generation problem into a multiple-choice problem.
\paragraph{Performance gain of {\verifyall}} While {\verifyall} helps LLMs obtain large performance gains in solving knowledge crosswords, it is quite intriguing when investigating where such gains come from. Specifically, in the \textsc{w/ knowledge} setting, among all 359 hard problems, we find only 3 problems whose solution with {\verifyall} involves detecting errors in verification and re-propose candidates. Among the 3 problems, 2 are answered correctly by both {\verifyall} and {\textsc{CoT}}, and both methods fail the other problem. This leads to an interesting implication that the performance gain comes from LLMs proposing more precise answers in the first attempt, and that LLMs can jointly consider all constraints rather than consider one by one. We envision the study of such performance gain and the application of the insight as possible future directions.
\paragraph{Application of geometric knowledge reasoning} Despite the difficulty of the task, LLMs do show preliminary geometric reasoning ability over incomplete fact network. While such ability still has a long way to achieve perfection, this finding opens up the possibility of using LLMs as flexible relational databases and accessing the parametric knowledge with prompts similar to SQL (structured query language).
\paragraph{Same prompting approach with different LLMs} While \textsc{gpt-3.5-turbo} does not benefit from {\stagebystage}, experiments using {\stagebystage} with \textsc{gpt-4} demonstrate impressive results under the \textsc{w/ knowledge} setting. Taking a close look at the responses of \textsc{gpt-3.5-turbo}, we find they fail to follow the reasoning steps presented in the exemplars even if we facilitate the process by guiding the \emph{update} step with program. On the other hand, \textsc{gpt-4} learn better from exemplars of {\stagebystage} with similar settings. This indicates that the success of {\stagebystage} relies heavily on the choice of LLMs.
\paragraph{Geometric Knowledge Reasoning vs.\ Logical Reasoning}
Logical reasoning, over natural language \citep{yang2022language,yang2023logical} or logical rules \citep{pan2023logic,luo2023chatrule}, could be confined within pre-defined logic operations set, while geometric knowledge reasoning problems are more flexible and involve diverse logical reasoning types, such as deductive reasoning (applying general knowledge and constraint patterns to deduce the correct answer), abductive reasoning (formulating the most likely answer based on the available clues), and more.
Considering the versatile nature of knowledge structure and flexible relational reasoning types involved, geometric knowledge reasoning is reasoning over natural language, without explicit transformation to logic rules as in logical reasoning.

\section{{\ourbenchmark} Details}
\label{sec:benchmark-details}
\begin{table}[t]
\centering
\begin{tabular}{@{}lc@{}}
\toprule[1.5pt]
\multicolumn{1}{c}{\textbf{Hyperparameter}} & \textbf{Value}      \\ \midrule[0.75pt]
$\text{degree}_c$                                       & 5, 7, 9             \\
$s_G$                                          & 6, 7, 8, 9, 10, 11  \\
$s_B$                                          & $[\frac{1}{4}\cdot s_G,\frac{1}{2}\cdot s_G]$ \\
$m_r$                                          & 1.1, 1.2, 1.3       \\
$m_b$                                          & 1, 1.1              \\ \bottomrule[1.5pt]
\end{tabular}
\caption{Hyperparamters for benchmark construction}
\label{tab:hyperparameter}
\end{table}

In this section, we elaborate on the details of benchmark construction and additional experiment details. %~\ref{sec:benchmark}
\subsection{Benchmark Construction Details}
\begin{enumerate}[leftmargin=*,itemsep=0pt,topsep=0pt,parsep=2pt,partopsep=0pt]
\item YAGO filtering
    \iitem We remove edges in YAGO with the following relations: (\romannumeral1) Location-related: isLocatedIn, livesIn,  happenedIn, diedIn, wasBornIn; (\romannumeral2) Time-sensitive: worksAt, playsFor, isAffiliatedTo, isPoliticianOf, isLeaderOf; (\romannumeral3) Not self-evident: influences, owns, isKnownFor, dealsWith, imports, exports, created, isInterestedIn, dealsWith, isConnectedTo.
    \iitem The remaining relations in YAGO are: graduatedFrom, hasMusicalRole, hasAcademicAdvisor, hasChild, wroteMusicFor, hasCapital, actedIn, hasOfficialLanguage, hasWonPrize, hasGender, hasCurrency, directed, isCitizenOf, participatedIn, isMarriedTo, hasNeighbor, edited.
\item Modified k-hop neighborhood
    \iitem A center node $c$ is randomly selected from nodes with degree $\text{degree}_c$ in filtered YAGO. 
    \iitem We retrieve a modified 5-hop neighborhood $\mathcal{G}_\mathcal{N}^{(c)}$: in each layer, we retain at most 8 nodes. This approach assists us in obtaining a subgraph with a relatively long diameter while avoiding excessive density.
\item Downsample to $\mathcal{G}_\mathcal{A}$
    \iitem We repeatedly remove nodes from $\mathcal{G}_\mathcal{N}^{(c)}$ until the number of nodes in the largest connected component in $\mathcal{G}_\mathcal{N}^{(c)}$ is no more than question graph size $s_G$. 
    \iiitem Sort the nodes in $\mathcal{G}_\mathcal{N}^{(c)}$ by degree in filtered YAGO in descending order as $\mathbf{v}_{\text{sorted},\text{YAGO}}$.
    \iiitem Denote \emph{reduce range multiplier} as $m_r$. Then \emph{reduce range} $rr$ is calculated as $m_r\cdot(|\mathcal{V}_{\mathcal{N}}^{(c)}|-s_G)$ where $\mathcal{V}_{\mathcal{N}}^{(c)}$ is the set of nodes in $\mathcal{G}_\mathcal{N}^{(c)}$. 
    \iiitem Randomly pick a node in $\mathbf{v}_{\text{sorted},\text{YAGO}}[(rr-1)/2:(rr-1)]$ and remove this node from $\mathcal{G}_\mathcal{N}^{(c)}$.
    \iitem Following the abovementioned approach, we downsample $\mathcal{G}_\mathcal{N}^{(c)}$ to $\mathcal{G}_\mathcal{A}$ by removing nodes with relatively high degree in filtered YAGO and introduce randomness in this process.
\item Generate blanks to get $\mathcal{G}_\mathcal{Q}$
    \iitem To mask $s_B$ nodes in $\mathcal{G}_\mathcal{A}$ as blanks, denote \emph{blank range multiplier} as $m_b$ and calculate \emph{blank range} $br$ as $s_B\cdot m_b$.
    \iitem Sort the nodes in $\mathcal{G}_\mathcal{N}^{(c)}$ by degree in $\mathcal{G}_\mathcal{Q}$ in descending order as $\mathbf{v}_{\text{sorted},\mathcal{G}_\mathcal{A}}$.
    \iitem We then randomly select $s_B$ nodes from the first $br$ nodes in $\mathbf{v}_{\text{sorted},\mathcal{G}_\mathcal{A}}$ as blanks.
\end{enumerate}
Specifically, the hyperparameters we used for benchmark construction are listed in Tabel~\ref{tab:hyperparameter}.
% Please add the following required packages to your document preamble:
% \usepackage{booktabs}
\subsection{Relevant Knowledge}
In the \textsc{w/ knowledge} setting, relevant knowledge is prepended to each problem. Specifically, for each triplet in the constraint, we present four pertinent triplets, with one of them reserved for the triplet containing the correct answer. 
And the other three are randomly sampled from YAGO following similar method as negative sampling, with priority given to the triplets that satisfy all the criteria (Rule \#1, \#2, and \#3).
By sampling relevant triplets in such a way, we provide necessary knowledge (filled with correct answers) as well as confounding knowledge that makes the solving process non-trivial. The sampling of confounding knowledge also simulates the possible information that one may consider when solving the question, with those satisfying all three criteria having the highest likelihood of being considered intuitively.
\subsection{Experiment Details}
Within 4k-context, in the w/o Relevant Knowledge setting, the numbers of finished responses for easy, medium, and hard questions are 755, 781, and 341 respectively; in the w/ Relevant Knowledge setting, the numbers of finished responses for easy, medium, hard questions are 759, 769 and 328 respectively. The credits are calculated based on these finished responses only.

% \newpage

\section{Additional Analysis}
\label{sec:add-analysis}

% \paravs

% \begin{figure}[t]
%     \centering
%     % \vspace{-0.13in}
    
%     \caption{Problem distribution based on the correctness under the \textsc{w/ knowledge} and \textsc{w/o knowledge} settings using \verifyall.}
%     \label{fig:error_verify}
%     % \vspace{-0.25in}
% \end{figure}

\begin{figure}[t]
    \centering
    % \vspace{-0.13in}
    \includegraphics[width=1\linewidth]{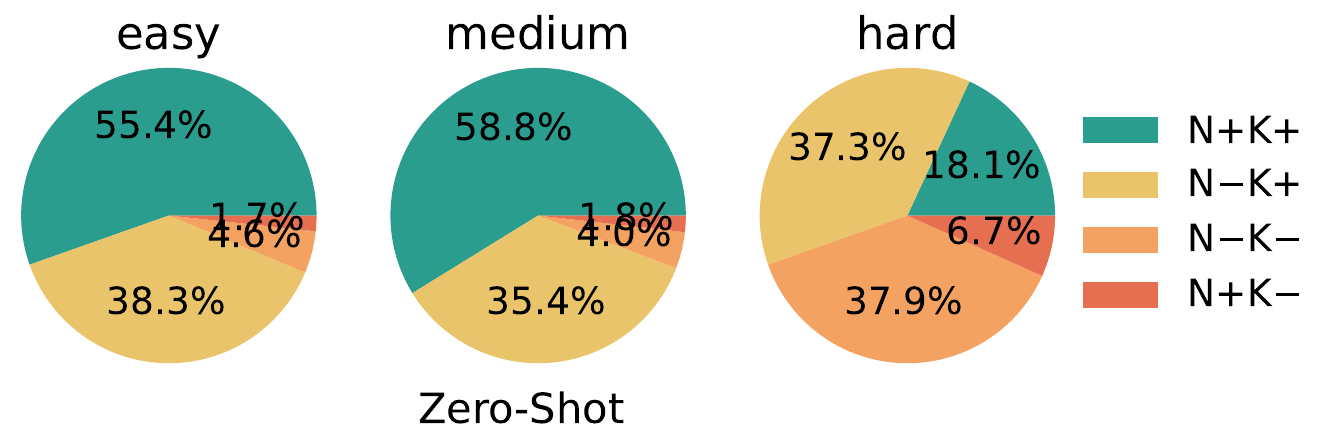}
    \caption{Problem distribution based on the correctness under the \textsc{w/ knowledge} and \textsc{w/o knowledge} settings using \textsc{Zero-Shot}.}
    \label{fig:error_zero}
    % \vspace{-0.25in}
\end{figure}

\paragraph{Error Analysis}
We provide error analysis conducted with \textsc{Zero-Shot} here for reference. Results in Figure \ref{fig:error_zero} shows that all three methods share similar trends.
% [[other pie charts]]

% \newpage

\paragraph{Number of Blanks}
We study the impact of the number of blanks in the problem on the model performance. Specifically, we randomly select 100 problems with three blanks (\# of blanks = 3) from all three difficulty levels and construct two additional versions of these problems by filling in one (\# of blanks = 2) or two (\# of blanks = 1) answers to the blanks. We evaluate the performance of various methods (\textsc{Zero-Shot}, \textsc{Few-Shot}, \textsc{CoT}, \textsc{Verify-All}) and two settings (\textsc{w/ knowledge} or \textsc{w/o knowledge}) on these three versions of the dataset and present the results in Table~\ref{tab:nblanks}.
We find that ChatGPT performs well on the simpler (\# of blanks = 1) version, demonstrating a strong knowledge ability. However, its performance suffers when the required reasoning steps increase and the geometric structures involved become more complex.

\begin{table*}
\centering
 \resizebox{1\linewidth}{!}{
\begin{tabular}{@{}clcccccccccccc@{}}
\toprule[1.5pt]
 & &
  \multicolumn{6}{c}{\textsc{w/ knowledge}} &
  \multicolumn{6}{c}{\textsc{w/o knowledge}} \\
  \cmidrule(lr){3-8} \cmidrule(lr){9-14}
\multirow{3}{*}{\textbf{\# of blanks}} & \multirow{3}{*}{\textbf{Methods}} &
  \multicolumn{2}{c}{\textbf{easy}} &
  \multicolumn{2}{c}{\textbf{medium}} &
  \multicolumn{2}{c}{\textbf{hard}} &
  \multicolumn{2}{c}{\textbf{easy}} &
  \multicolumn{2}{c}{\textbf{medium}} &
  \multicolumn{2}{c}{\textbf{hard}} \\
  \cmidrule(lr){3-4} \cmidrule(lr){5-6}  \cmidrule(lr){7-8}  \cmidrule(lr){9-10}  \cmidrule(lr){11-12}  \cmidrule(lr){13-14}
  & &
  \textsc{PC} &
  \textsc{FC} &
  \textsc{PC} &
  \textsc{FC} &
  \textsc{PC} &
  \textsc{FC} &
  \textsc{PC} &
  \textsc{FC} &
  \textsc{PC} &
  \textsc{FC} &
  \textsc{PC} &
  \textsc{FC} \\ \midrule[0.75pt]
\textsc{3} &\textsc{Zero-Shot} 
& 98.0         & 95.0         & 98.3         & 95.0         & 77.7          & 51.0          & 85.0          & 64.0          & 87.3          & 68.0          & 59.3 & 21.0 \\
3 & \textsc{Few-Shot} & 98.7 & 96.0 & 98.0 & 95.0 & 77.3 & 44.0 & 83.0 & 57.0 & 84.0 & 61.0 & 52.0 & 12.0 \\
3 & \textsc{CoT} & 93.7 & 87.0 & 94.3 & 90.0 & 78.0 & 52.0 & 73.7 & 45.0 & 79.3 & 49.0 & 49.3 & 18.0 \\
3 & \textsc{Verify-All} & 99.3 & 98.0 & 98.0 & 94.0 & 81.0 & 54.0 & 86.3 & 64.0 & 88.7 & 68.0 & 57.0 & 21.0 \\
2 & \textsc{Zero-Shot} & 96.0 & 94.0 & 93.5 & 90.0 & 78.0 & 67.0 & 84.5 & 73.0 & 82.5 & 68.0 & 61.0 & 34.0 \\
2 & \textsc{Few-Shot} & 97.5 & 95.0 & 96.5 & 93.0 & 85.5 & 74.0 & 82.5 & 68.0 & 86.0 & 74.0 & 66.0 & 44.0 \\
2 & \textsc{CoT} & 91.5 & 85.0 & 93.0 & 89.0 & 82.0 & 71.0 & 78.5 & 62.0 & 78.5 & 64.0 & 58.5 & 41.0 \\
2 & \textsc{Verify-All} & 98.5 & 97.0 & 98.0 & 96.0 & 89.0 & 81.0 & 87.0 & 79.0 & 87.0 & 76.0 & 60.5 & 45.0 \\
1 & \textsc{Zero-Shot} & 96.0 & 96.0 & 91.0 & 91.0 & 76.0 & 76.0 & 92.0 & 92.0 & 88.0 & 88.0 & 63.0 & 63.0 \\
1 & \textsc{Few-Shot} & 99.0 & 99.0 & 97.0 & 97.0 & 87.0 & 87.0 & 92.0 & 92.0 & 92.0 & 92.0 & 64.0 & 64.0 \\
1 & \textsc{CoT} & 97.0 & 97.0 & 93.0 & 93.0 & 83.0 & 83.0 & 92.0 & 92.0 & 90.0 & 90.0 & 59.0 & 59.0 \\
1 & \textsc{Verify-All} & 100.0 & 100.0 & 97.0 & 97.0 & 91.0 & 91.0 & 92.0 & 92.0 & 87.0 & 87.0 & 69.0 & 69.0\\

  \bottomrule[1.5pt]
\end{tabular}

}

\caption{FC and PC with \textsc{gpt-3.5-turbo} on 100 randomly sampled problems with three blanks. We fill in (3 - \# of blanks) blanks in each problem and the LM is tasked with figuring out the remaining blanks.
% [one sentence about how verify-all outperforms baselines by at least ?\% on hard subsets]
}
\label{tab:nblanks}
\end{table*}

\section{Qualitative analysis}
In this section, we provide examples of knowledge crosswords that \textsc{gpt-3.5-turbo} answers correctly or wrongly using {\stagebystage} and {\verifyall}. In-context exemplars are omitted in this section to save space and can be found in Appendix \ref{sec:appendix-prompts}. Table~\ref{tab:stage-correct} and Table~\ref{tab:stage-wrong} show results using {\stagebystage}; Table~\ref{tab:verifyall-correct-backtracking}, Table~\ref{tab:verifyall-correct-single}  and Table~\ref{tab:verifyall-wrong} show results using {\verifyall}.

\vspace{13pt}
\section{Prompts}
\label{sec:appendix-prompts}
We list the prompts for all experiments of Tables~\ref{tab:main-results} and~\ref{tab:finetune} in Tables~\ref{tab:promp-ub0shot},~\ref{tab:promp-few},~\ref{tab:promp-cot},~\ref{tab:promp-ltm},~\ref{tab:promp-stage1},~\ref{tab:promp-stage2},~\ref{tab:promp-verify}.
% \newpage
\begin{table*}[t]
\caption{Response using {\stagebystage} where the answers are correct.}
\begin{adjustbox}{max width=1\textwidth}
\begin{tabular}{p{4in}p{5in}}
    \toprule[1.5pt]
    
    {\textbf{Prompt}}   & {\textbf{Response}}  \\
\end{tabular}
\end{adjustbox}
\label{tab:stage-correct}
\end{table*}

% \newpage
\begin{table*}[t]
\caption{Response using {\stagebystage} where the answers are wrong.}
\begin{adjustbox}{max width=1\textwidth}
\begin{tabular}{p{4in}p{5in}}
    \toprule[1.5pt]
    
    {\textbf{Prompt}}   & {\textbf{Response}}  \\
\end{tabular}
\end{adjustbox}
\label{tab:stage-wrong}
\end{table*}
 
\newpage
\begin{table*}[t]
\caption{Response using {\verifyall} where the answers are correct and involves error detection and backtracking.}
\begin{adjustbox}{max width=1\textwidth}
\begin{tabular}{p{5.5in}p{3.5in}}
    \toprule[1.5pt]
    
    {\textbf{Prompt}}   & {\textbf{Response}}  \\
\end{tabular}
\end{adjustbox}
\label{tab:verifyall-correct-backtracking}
\end{table*}

\begin{table*}[t]
 \caption{Response using {\verifyall} where the answers are correct correctly in a single trial.}
\begin{adjustbox}{max width=1\textwidth}
\begin{tabular}{p{5.5in}p{3.5in}}
    \toprule[1.5pt]
    
    {\textbf{Prompt}}   & {\textbf{Response}}  \\
         \midrule[1pt]
    % \tiny 
     [{\verifyall} exemplars]
     \quad

    Instruction: Pick the correct answer for each blank that satisfies all the given constraints.
    
    Knowledge: (Paz Vega, acted in, The Human Contract); (Joe Roberts, acted in, Our Hospitality); (Paz Vega, acted in, The Six Wives of Henry Lefay); (Paz Vega, acted in, The Spirit (film)); (Charlie Chaplin, directed, A Woman of Paris); (Charlie Chaplin, directed, The Champion (1915 film)); (Charlie Chaplin, directed, Limelight (1952 film)); (Jada Pinkett Smith, directed, The Human Contract); (Johnathon Schaech, acted in, Prom Night (2008 film)); (Joe Roberts, acted in, Our Hospitality); (Brittany Snow, acted in, Prom Night (2008 film)); (Idris Elba, acted in, Prom Night (2008 film)); (Joe Roberts, acted in, The Haunted House (1921 film)); (Idris Elba, acted in, The Human Contract); (Joe Roberts, acted in, Our Hospitality); (William Hopper, acted in, The Bad Seed (1956 film)).
    
    Constraints: (Paz Vega, acted in, blank 1); (Jada Pinkett Smith, directed, blank 1); (blank 2, acted in, Prom Night (2008 film)); (blank 2, acted in, blank 1).
    
    Options:
    blank 1: A. The Human Contract, B. The Spirit (film), C. The Six Wives of Henry Lefay
    blank 2: A. Johnathon Schaech, B. Idris Elba, C. Brittany Snow
    
    Answer:

    & 
    Candidate answer: blank 1: A. The Human Contract, blank 2: B. Idris Elba;
    
    Filled constraints with candidate answer: (Paz Vega, acted in, The Human Contract); (Jada Pinkett Smith, directed, The Human Contract); (Idris Elba, acted in, Prom Night (2008 film)); (Idris Elba, acted in, The Human Contract);
    
    Verification: Does error occur in filled constraints with candidate answer?
    No.
    Therefore, blank 1: A, blank 2: B.
    \sethlcolor{yellow}\hl{\checkmark}
    \\ \bottomrule[1.5pt]
\end{tabular}
\end{adjustbox}
\label{tab:verifyall-correct-single}
\end{table*}

\begin{table*}[t]
\caption{Response using {\verifyall} where the answers are wrong.}
\begin{adjustbox}{max width=1\textwidth}
\begin{tabular}{p{5.5in}p{3.5in}}
    \toprule[1.5pt]
    
    {\textbf{Prompt}}   & {\textbf{Response}}  \\
         \midrule[1pt]
    % \tiny 
      [{\verifyall} exemplars]
     \quad

     Instruction: Pick the correct answer for each blank that satisfies all the given constraints.
     
    Knowledge: (Colleen Camp, has gender, female); (Robert Bly, has gender, male); (Eric Maskin, has gender, male); (Charlie Chaplin, has gender, male); (Grace Zabriskie, has gender, female); (Robert Bly, has gender, male); (Eric Maskin, has gender, male); (Charlie Chaplin, has gender, male); (Tim Conway, acted in, The Private Eyes (1980 film)); (Don Knotts, acted in, The Private Eyes (1980 film)); (Bernard Fox (actor), acted in, The Private Eyes (1980 film)); (Grace Zabriskie, acted in, The Private Eyes (1980 film)); (Charlie Chaplin, has gender, male); (Eric Maskin, has gender, male); (Diane Baker, has gender, female); (Robert Bly, has gender, male); (Jane Seymour, has gender, female); (Eric Maskin, has gender, male); (Robert Bly, has gender, male); (Charlie Chaplin, has gender, male).
    Constraints: (Colleen Camp, has gender, blank 2); (blank 1, has gender, blank 2); (blank 1, acted in, The Private Eyes (1980 film)); (Diane Baker, has gender, blank 2); (Jane Seymour, has gender, blank 2).
    Options:
    blank 1: A. Bernard Fox (actor), B. Grace Zabriskie, C. Don Knotts
    blank 2: A. female, B. male
    Answer:

    & 
    Candidate answer: blank 1: C. Don Knotts, blank 2: A. female;
    
    Filled constraints with candidate answer: (Colleen Camp, has gender, female); (Don Knotts, has gender, female); (Don Knotts, acted in, The Private Eyes (1980 film)); (Diane Baker, has gender, female); (Jane Seymour, has gender, female);
    
    Verification: Does error occur in filled constraints with candidate answer?
    No.
    Therefore, blank 1: C, blank 2: A.
    \sethlcolor{yellow}\hl{\XSolidBrush~(Correct answer: blank 1: B, blank 2: A)}
    \\
\end{tabular}
\end{adjustbox}
\label{tab:verifyall-wrong}
\end{table*}

% \newpage

%Upperbound, Zero-Shot
\begin{table*}[t]
\caption{Prompts used in main experiments with exemplars and an example question. ``Knowledge'' for each problem is only applicable in the ``\textsc{w/ knowledge}'' setting.}
\label{tab:promp-ub0shot}
\begin{adjustbox}{max width=1\textwidth}
%
\end{adjustbox}
\end{table*}

% few shot
\begin{table*}[t]
\caption{Prompts used in main experiments with exemplars and an example question. ``Knowledge'' for each problem is only applicable in the ``\textsc{w/ knowledge}'' setting.}
\label{tab:promp-few}
\begin{adjustbox}{max width=1\textwidth}
%
\end{adjustbox}
\end{table*}

% CoT 
\begin{table*}[t]
\caption{Prompts used in main experiments with exemplars and an example question. ``Knowledge'' for each problem is only applicable in the ``\textsc{w/ knowledge}'' setting.}
\label{tab:promp-cot}
\begin{adjustbox}{max width=1\textwidth}
%
\end{adjustbox}
\end{table*}

% LtM
\begin{table*}[t]
\caption{Prompts used in main experiments with exemplars and an example question. ``Knowledge'' for each problem is only applicable in the ``\textsc{w/ knowledge}'' setting.}
\label{tab:promp-ltm}
\begin{adjustbox}{max width=1\textwidth}
%
\end{adjustbox}
\end{table*}

% stage by stage
\begin{table*}[t]
\caption{Prompts used in main experiments with exemplars and an example question. ``Knowledge'' for each problem is only applicable in the ``\textsc{w/ knowledge}'' setting.}
\label{tab:promp-stage1}
\begin{adjustbox}{max width=1\textwidth}
%
\end{adjustbox}
\end{table*}
% stage by stage
\begin{table*}[t]
\caption{Prompts used in main experiments with exemplars and an example question. ``Knowledge'' for each problem is only applicable in the ``\textsc{w/ knowledge}'' setting.}
\label{tab:promp-stage2}
\begin{adjustbox}{max width=1\textwidth}
%
\end{adjustbox}
\end{table*}

% verify-all
\begin{table*}[t]
\caption{Prompts used in main experiments with exemplars and an example question. ``Knowledge'' for each problem is only applicable in the ``\textsc{w/ knowledge}'' setting.}
\label{tab:promp-verify}
\begin{adjustbox}{max width=1\textwidth}
%
    \end{adjustbox}
    \\ \bottomrule[1.5pt]
\end{tabular}
\end{adjustbox}
\end{table*}

\end{document}